\documentclass[runningheads]{llncs}

\usepackage{eccv}

\usepackage{xspace}
\usepackage{bm}
\usepackage{overpic}
\usepackage{pgfplots}
\pgfplotsset{compat=1.18}
\usepgfplotslibrary{fillbetween}
\usepackage{tikz}
\usetikzlibrary{positioning}
\usetikzlibrary{pgfplots.groupplots}
\usetikzlibrary{calc}
\usetikzlibrary{shapes}
\usetikzlibrary{backgrounds}

\usepackage{siunitx}
\usepackage{booktabs}
\usepackage{wrapfig}
\usepackage{float}

\newcommand{\repname}{FUSE\xspace}

\newcommand{\R}{\mathbb R}

\usepackage{multirow}
\usepackage{booktabs}
\usepackage{multirow}
\usepackage{graphicx}

\usepackage{eccvabbrv}

\usepackage{graphicx}
\usepackage{booktabs}

\usepackage[accsupp]{axessibility}  % Improves PDF readability for those with disabilities.

\usepackage{hyperref}

\usepackage{orcidlink}

\begin{document}

% ---------------------------------------------------------------
\title{FUSE: A Flow-based Mapping Between Shapes}

\author{Lorenzo Olearo\inst{1}\orcidlink{0009-0000-7290-3549} \and
Giulio Viganò\inst{1}\orcidlink{0009-0000-7263-2160} \and
Filippo Maggioli \inst{4}\orcidlink{0000-0001-8008-8468} \and
Daniele Baieri \inst{1, 2, 3}\orcidlink{0000-0002-0704-5960} \and
Simone Melzi \inst{1}\orcidlink{0000-0003-2790-9591}}

\authorrunning{L.~Olearo et al.}

\institute{
  University of Milano-Bicocca,
  %Department of Informatics, Systems and Communication (DISCo),
  Milan, Italy
  %\email{\{lorenzo.olearo, giulio.vigano, simone.melzi\}@unimib.it}
  \and
  University of Bonn, 
  %Department of Computer Science, 
  Bonn, Germany\\
  %\email{dbaieri@uni-bonn.de}
  \and
  Lamarr Institute, Germany
  \and
  Pegaso University, 
  %Department of Information Science and Technology,
  Naples, Italy\\
  %\email{maggioli.filippo@gmail.com}
}

\maketitle
\input{Figures/teaser}

\begin{abstract}
We introduce a novel neural representation for maps between 3D shapes based
on flow-matching models, which is computationally efficient and supports
cross-representation shape matching without large-scale training or
data-driven procedures. 3D shapes are represented as the probability
distribution induced by a continuous and invertible flow mapping from a fixed
anchor distribution. Given a source and a target shape, the composition of
the inverse flow (source to anchor) with the forward flow (anchor to target),
we map points between the two surfaces.  By encoding the shapes with a
pointwise task-tailored embedding, this construction provides an invertible
and modality-agnostic representation of maps between shapes across point
clouds, meshes, signed distance fields (SDFs), and volumetric data. The
resulting representation consistently achieves high coverage and accuracy
across diverse benchmarks and challenging settings in shape matching. Beyond
shape matching, our framework shows promising results in other tasks,
including UV mapping and registration of raw point cloud scans of human
bodies.
% We introduce a novel neural representation for maps between 3D shapes based
%     on flow-matching models, computationally efficient and supporting
%     cross-representation shape matching without large-scale training or
%     data-driven procedures. 3D shapes are represented as the probability
%     distribution induced by a continuous, invertible flow from a fixed anchor.
%     Given source and target shapes, we map points between the two surfaces by
%     composing the inverse flow (source to anchor) with the forward flow (anchor
%     to target). Encoding shapes with a pointwise task-tailored embedding, this
%     construction yields an invertible, modality-agnostic representation of maps
%     across point clouds, meshes, signed distance fields (SDFs), and volumetric
%     data. The representation consistently achieves high coverage and accuracy
%     across diverse benchmarks and challenging settings. Beyond matching, our framework shows promising results on UV mapping and
%     registration of raw human body scans.
\end{abstract}

\section{Introduction}
    
%This paper introduces a new framework that blends flow matching theory and optimal transport for accurate, consistent, and flexible shape matching.

%% Daniele: I'm being really brief on purpose for now, feel free to add something if you want. Especially citations.
Shape matching has long been recognized as a ``key problem'' in geometry
processing: indeed, accurate correspondences between two shapes enable simple
solutions for several downstream tasks (\eg, pose transfer
\cite{song20213dposetransfercorrespondence} or interpolation
\cite{eisenberger2021neuromorphunsupervisedshapeinterpolation}). Most of the
research in this field, following the common practices from 3D graphics
industry, was built on the triangle mesh representation, which approximates a
continuous 2-dimensional manifold embedded in $\R^3$ piecewise linearly with a
collection of triangles connected by their common edges and vertices. However,
recent progress in the fields of computer vision and deep learning has
popularized novel ways of encoding 3D geometry, such as neural representations
\cite{park2019deepsdf, mildenhall2020nerf}. These solutions offer several
advantages compared to triangle meshes, but lack their large suite of tools and
algorithms for analysis and manipulation \cite{nfgp}.

In this work, we address this challenge by introducing a novel representation for maps between a source and a target shape that can be applied over various geometric encodings. We namely refer to this representation as \repname. Constructing a map with this representation requires only two components: i) the ability to sample points from the involved surfaces, and ii) a pointwise feature embedding designed for the target task.
Once feature embeddings are computed for the sampled points on both source and target shapes, \repname estimates an invertible flow from each embedding distribution to a shared anchor distribution, following Flow Matching framework \cite{lipman2022flow}. \repname then expresses the map between the two shapes as the composition of the inverse flow of the source (source $\to$ anchor) and the forward flow of the target (anchor $\to$ target).
%Note that, given the flows of any two shapes, the map is obtained in a ''zero-shot'' fashion, \ie, without any additional optimization or data.
We note that, given the flows of any two shapes, the map is obtained with no additional optimization, with only landmarks as information shared by the two shapes.

In the context of shape matching, our representation introduces a probabilistic perspective on embedding alignment, where semantically similar features are encouraged to align through a shared probability distribution. Combined with the guarantee of mapping points directly into the target embedding space, our framework produces high-coverage and accurate correspondences that are relevant for different applications.
% In the context of shape matching, our representation introduces a novel probabilistic perspective on embedding alignment, pushing semantically similar encodings into the same probability distribution. We introduce a probabilistic prior that makes embedding more aligned. This, combined with the guarantee of targeting points in the target embedding space, our pipeline returns high coverage maps useful for diverse applications. 
We evaluate the proposed approach through a comprehensive experimental study, comparing it against several shape matching baselines, ranging from well-established axiomatic methods to recent data-driven approaches; \repname shows high, if not the highest, accuracy on various benchmarks.
% Finally, we study the properties of the resulting shape matching method via a comprehensive evaluation, comparing FUSE matching with multiple baselines, ranging from classic algorithms to recent contributions with very high accuracy.
Our contributions can be summarized as follows:
\begin{itemize}
\item We introduce \repname, the first flow-matching based representation for 3D shape maps that naturally enables correspondence between shapes expressed in different modalities (e.g., meshes, point clouds, and SDFs);
\item We quantitatively show that \repname achieves competitive or superior matching accuracy over state-of-the-art methods, particularly in the context of landmark-based shape matching;
\item We showcase the versatility of \repname, employing it as a backbone for other downstream applications, ranging from shape parametrization to point cloud deformation.
\end{itemize}
% Summarizing, the contributions of our work are the following:
% \begin{itemize}
%     \item We introduce \repname, the first neural shape representation which naturally enables computing zero-shot correspondences between shapes in different representations (\eg, mesh with SDF);
%     \item \sm{check if we should tone down the claim} We show that \repname provides an improvement in matching accuracy over the state of the art, especially for axiomatic methods. %% Daniele: this still has to be confirmed
%     \item We employ \repname as a backbone matching model for downstream applications such as \daniele{texture and parametrization transfer on neural fields and point cloud deformation.}
% \end{itemize}

\section{Related Work}
%Our method combines classical approaches to shape correspondence with novel frameworks for flow matching in shape representation.

\subsection{Neural Shape Representations}
Recent research in geometric deep learning and geometry processing popularized
the practice of encoding 3D shapes in the weights of neural networks, which is
referred to as neural shape representation. The early contributions in this
field are due to \cite{OccupancyNetworks} and \cite{park2019deepsdf}, followed
by the introduction of NeRF~\cite{mildenhall2020nerf}, which led to a vast
effort to adapt neural shape representations to other tasks besides
reconstruction. Examples of these adaptations are to shape
deformation~\cite{implicitarap,nfgp,Novello_2023_ICCV,njf}, text-guided shape
editing~\cite{editnerf,Wang2021CLIPNeRFTD} and
generation~\cite{poole2023dreamfusion,Liu_2023_ICCV}. Recent contributions also
introduced neural representations for 3D simulations~\cite{pan22nif} and
computation of geometric operators over spherical neural
surfaces~\cite{williamson25sns}.

The work of~\cite{koestler22intrinsicneuralfields} is closely related to ours,
as they propose a method to represent functions continuously over surfaces in
intrinsic space. 
%\sm{The following is a fundamental sentence, will be important to state already here why are we different} 
Lastly, \repname's architecture is inspired by Geometry
Distributions~\cite{zhang2024geometry}, which proposes to represent geometry as
a continuous map from a 3D Gaussian distribution to the surface with a denoising
diffusion model \cite{song2022denoisingdiffusionimplicitmodels}. However, by
targeting the specific goal of representing maps between shapes, we build upon
flow matching~\cite{lipman2022flow},
%\sm{dovremmo prima di tutto introdurre il lavoro lipman2022flow e magari anche motivare la nostra scelta}
%\giulio{aggiunto lipman in introduction} 
and model flows directly over an intrinsic feature space, rather than standard Euclidean or extrinsic coordinates.
These two choices together enable \repname to represent mappings, as we prove by targeting the shape correspondence task.

%\sm{Aggiungere una frase che rimandi alla subsection sguente, dicendo che un esempio, se non l'esempio principale, di mappe che vogliamo rappresentare sono corrispondenze.}

\subsection{Shape Matching}
Shape correspondence is a fundamental task in geometry processing, with a wide range of approaches proposed over the years, which can largely be distinguished by how they represent the map between shapes and features~\cite{sun2009concise, aubry2011wave,shot,WFT,bestpaper}.
Among these, the functional map framework~\cite{ovsjanikov2012functional} and its follow-up works~\cite{melzi2019zoomout, marin2020correspondence,donati2022deep,cao23unsup,maggioli2025rematching} have emerged as a dominant paradigm, where correspondences are represented as linear operators between functional spaces. These methods offer attractive properties such as smoothness and invariance to deformations, but the reliance on spectral bases limits their generalization to non-mesh representations~\cite{marin2020correspondence} and in obtaining bijective maps~\cite{marin2020cmh, bastian2024hybridfunctionalmapscreaseaware, vigano2024bule}.

Another class of approaches leverages optimal transport (OT)~\cite{cuturi2013sinkhorndistanceslightspeedcomputation,liu2020semantic,saleh2022bendinggraphshierarchicalshape}, in which correspondences are represented by transport plans that minimize the cost of transporting a distribution of features from one shape to another. In the presence of sampling imbalances or ambiguous features, OT plans often exhibit mass-splitting behavior, where individual points distribute probability mass across multiple targets. This diffusion can degrade correspondence sharpness and stability when recovering discrete maps.
%When both shapes are discretized with equal numbers of uniformly weighted samples, OT yields a permutation matrix that defines a global bijection between points. However, this assumption becomes problematic when comparing shapes with different mesh densities or feature representations, since the mass-preserving constraint no longer holds.
Alternatively, registration-based approaches directly optimize non-linear deformation fields in 3D~\cite{li2022nonrigid, trappolini2021shape, raganatoetalSGP2023, eisenberger2021neuromorphunsupervisedshapeinterpolation}. The fields are often parameterized via deformation networks or complex architectures, achieving accurate alignments at the cost of solving computationally heavy optimization problems and depending on rigid initialization or learned priors.

Hybrid approaches combine the strengths of the above formulations~\cite{le2024integratingefficientoptimaltransport, pai2021sikhorn, jiang2023nonrigidshaperegistrationdeep, cao23unsup, vigano2025nam, cao2024spectralmeetsspatialharmonising, hartman2023bareesariemannianframeworkunregistered}, gaining in efficiency and robustness, but still relying on learned priors and struggling to generalize to other representations. Another common approach is template-based matching, representing correspondences as a composition of maps with respect to a common template, selected a priori \cite{groueix20183dcoded, marin2024nicpneuralicp3d, FARM} or directly at test time \cite{magnet:2023:craniofacial}.
On a different perspective, \repname offers a novel representation for maps. Based on flow matching ~\cite{lipman2022flow}, our method models a correspondence as a composition of continuous invertible flows between intrinsic feature spaces and a shared anchor distribution, inherently enforcing bijectivity and high-coverage in obtained maps, without learned priors or per-pair optimization in a pipeline naturally agnostic of the geometric representation.

% Hybrid approaches have emerged that aim to combine the strengths of these
% methods, e.g., integrating functional maps with OT solvers or learning
% feature-aligned
% transformations
% ~\cite{le2024integratingefficientoptimaltransport, pai2021sikhorn, jiang2023nonrigidshaperegistrationdeep}.

%Finally, other, less related methods avoid explicit feature computation altogether. These include solving combinatorial matching thrugh linear programming \cite{roetzer2025fastgloballyoptimalgeometrically}, using neural implicit fields \cite{marin2024nicpneuralicp3d}, or relying on attention mechanisms for geometric learning \cite{raganatoetalSGP2023, riva2024localizedgaussiansselfattentionweights}.

\section{Background}

% In this section, we introduce the necessary notation and context necessary to outline our method.

\paragraph{Shapes and Matching.}
A 3D shape $\mathcal{S}$ is modeled as a compact two-dimensional manifold
embedded in $\mathbb{R}^3$. From a computational standpoint, single shapes may be represented in several ways: polygonal meshes, point clouds, signed distance functions and volumetric meshes are notable examples. In this work, we identify a shape $\mathcal{S}$ as a continuous
manifold from which we can sample unordered sets of points
$\{x_i\}_{i=0}^{n}$, encoded by their 3D coordinates (i.e. $x_i \in \mathbb{R}^3$) independently from the
adopted representations. 

Given 3D shapes $\mathcal{S}_1=\{x_{i}\}_{i=1}^{n_1}$ and
$\mathcal{S}_2=\{y_{j}\}_{j=1}^{n_2}$ with a common semantic, a point-to-point
correspondence $T_{12}$ maps each point $x_i$ to its corresponding point
$y_j$. Shape matching is the task of estimating such correspondence. 
% $T_{12}$ can be encoded as the matrix $\Pi_{21}\in \mathbb{R}^{n_1 \times n_2} $ such that $\Pi_{21}(i,j)=1$, if $T_{12}(x_i)=y_j$ and $\Pi_{21}(i,j)=0$ otherwise.  If the correspondence is bijective, $\Pi_{21}$ is a permutation matrix.
%\paragraph{Feature matching}
In the non-rigid scenario, estimating correspondences directly from 3D coordindates is challenging;
% it is difficult to estimate the correspondence working directly with the 3D coordinates of points; 
thus, most methods describe the shapes via pointwise \emph{features}, characterizing their semantics and geometry~\cite{sun2009concise,shot,Dutt_2024_CVPR}. For
each point $x_i \in \mathcal{S}_1$, we compute a vector of features $f_i\in
\mathbb{R}^{d}$ that we can then store as a row in the \emph{embedding matrix} $E_1\in
\mathbb{R}^{n_1\times d}$ (same goes for $\mathcal{S}_2$). %, which contains the embedding of all the points sampled from $\mathcal{S}_1$ (similarly for $\mathcal{S}_2$). 
Ideally, one can solve for
the correspondence directly by leveraging the similarity between these encodings~\cite{aubry2011wave}. In practice, 
%it is necessary to design a shape-matching pipeline to improve the alignment of 
more complex pipelines are required to align the features and retrieve the correspondence from their relations~\cite{ovsjanikov2012functional}.

\input{Figures/pipeline_3d}

\paragraph{Flow Matching.}
Flow Matching (FM)~\cite{lipman2022flow, lipman2024flow} is a framework for
learning continuous in time probability paths $(p_t)_{t \in [0,1]}$ that transport
a known source distribution $p_0$ (\eg, a Gaussian) to an unknown target $p_1$
(\eg, data samples). This is achieved by learning a \emph{velocity field}
$u_t^\theta(x)$ describing the instantaneous dynamics of the path, parameterized by a
neural network $\theta$.  
For each $x \in \mathbb{R}^d$, the velocity field induces a flow $\psi_t(x)$ as
the solution of the ODE $\frac{d}{dt}\psi_t(x) = u_t(\psi_t(x))$ with initial
condition $\psi_0(x) = x$. Crucially, the flow $\psi_t$ is invertible and
diffeomorphic: starting from any point sampled from $p_1$, i.e., $x \sim p_1$,
integrating the ODE backward yields a point $\psi^{-1}_0(x) \sim p_0$. The
bidirectionality enables transport between distributions in both
directions.  
% A naive training objective would regress a network $v^\theta_t(x)$ on the
% unknown velocity $u_t(x)$:
% \begin{equation}
% \mathcal{L}_{\text{FM}}(\theta) = \mathbb{E}_{t, x_t \sim p_t} \left[ \lVert v^\theta_t(x_t) - u_t(x_t)\rVert^2 \right].
% \end{equation}
% where $\mathbb{E}_{t, x_t \sim p_t} $ represents the average over randomly sampled timesteps in $[0,1]$ and points along the path.
Since $u_t$ is generally intractable,~\cite{lipman2022flow,lipman2024flow}
proposed Conditional Flow Matching (CFM), replacing $u_t$ with a tractable
conditional velocity field $u_t(x_t \mid x_1)$, while preserving the same
gradients:
\begin{equation}
\mathcal{L}_{\text{CFM}}(\theta) = \mathbb{E}_{t, x_1, x_t \sim p_t(\cdot|x_1)} \left[ \lVert v^\theta_t(x_t) - u_t(x_t \mid x_1)\rVert^2 \right].
\end{equation}
where $\mathbb{E}_{t, x_1, x_0}$ represents the average over randomly sampled
timesteps in $[0,1]$ and over source and target  samples $x_0 \sim p_0,x_1 \sim
p_1$. A simple choice for the field is linear interpolation, $x_t = (1-t)x_0 + t
x_1$ with $u_t = x_1 - x_0$. CFM training is thus simulation-free, although
generating samples still requires integrating the learned ODE.
In the remainder of the paper, we will refer to CFM simply as Flow Matching (FM).

\section{\repname}\label{sec:method}

Given the overview presented in the previous section, we can now provide all the
details of our novel representation \repname which we outlined in
\cref{fig:teaser}. The complete source code is available at
\url{https://github.com/LorenzoOlearo/FUSE-Flow-based-Mapping-Between-Shapes}.

\subsection{Flow Optimization}\label{sec:flowopt} Given a pair of shapes
$\mathcal{S}_1=\{x_{i}\}_{i=1}^{n_1}$, $\mathcal{S}_2=\{y_{j}\}_{j=1}^{n_2}$ and
an embedding matrix $E_\ell\in \mathbb{R}^{n_\ell\times d}$ (with $\ell =1$ or
$2$), we denote the resulting embedding distributions as $p^{\ell} = E_{\ell}$
(see blue box of~\cref{fig:teaser}).
For each shape, we train an independent flow model that learns a continuous,
invertible mapping $\psi^{\ell}\colon[0,1]\times\mathbb{R}^d\to\mathbb{R}^d$.
This flow transports samples from a fixed, shared Gaussian $p_0 =
\mathcal{N}(0,1)^d$ (the anchor distribution) to the shape's embedding
distribution $p^{\ell}$, i.e. $ \psi^{\ell}_1(x) \sim p^{\ell}$ given $x\sim p_0$. Each
flow is parametrized by a neural network and trained via Conditional Flow
Matching~\cite{lipman2022flow, lipman2024flow} as we described earlier (see red
box of~\cref{fig:teaser}).
Each flow is trained independently on its shape, requiring only the ability to
sample embeddings from the surface.

\input{Figures/pipeline_encoding}

\subsection{Fusing Representations}\label{sec:bridge}
\repname determines maps between shapes by composing their flows. Given the two
trained flows $\psi^{1}$ and $\psi^{2}$, we define the map from $\mathcal{S}_1$
to $\mathcal{S}_2$ as:
\begin{equation}
\Phi^{12} = \psi^{2} \circ (\psi^{1})^{-1}
\end{equation}
This composition leverages the Gaussian distribution as a common anchor: we
transport the embedding points from $\mathcal{S}_1$ back to the Gaussian via
$(\psi^{1})^{-1}$, then forward to $\mathcal{S}_2$ via $\psi^{2}$ as depicted in
~\cref{fig:shape-interpolation}.
This process offers several advantages:
\begin{enumerate}
    \item Injection of invariance of the adopted embeddings in the output map;
    \item For any pair of shapes represented by the respective
    flows from a common distribution, no extra training is required to
    obtain a map between them;
    \item Representation invariance: creating a bridge with \repname only
    requires sampling from the surface, which is straightforward on most 3D
    representations.
\end{enumerate}
We can leverage this representation of maps between shapes to perform various tasks on the discrete geometry. For instance, to retrieve the correspondence between points of two discretized shapes, we can perform a nearest neighbor search in the embedding space.
\label{eq:knn}
\begin{equation}
    T_{12} = \text{NearestSearch}(\Phi^{12} (E(S_1)), E(S_2)) 
\end{equation}
%

%Given the nature of the FM solutions and the use of a shared distribution as an
%anchor, points with similar embeddings from different shapes are mapped to
%nearby regions within this anchor (see the green box in \cref{fig:teaser}). This
%relies on the assumption that both shapes share the same underlying semantic
%structure, encoded by the sets of pointwise embeddings.

%%% GIULIO's VERSION %%%
The validity of this construction rests on the nature of the FM solutions and
the shared anchor distribution. While FM training relies on independent random
couplings between Gaussian and data samples, this randomness affects only the
conditional velocity fields, not the learned flow. Theorem~2
in~\cite{lipman2022flow} shows that marginalizing the conditional fields $u_t(x
\mid x_1)$ over $x_1 \sim p_1$ recovers the \emph{unique} marginal field $u_t$
generating the probability path between $p_1$ and $p_0$; since this field is
determined by $(p_1, p_0)$ alone, so is the induced flow $\psi_t$. Consequently,
if $E(S_1) = E(S_2) = p_1$, the flows $\psi^1$ and $\psi^2$ coincide regardless
of the couplings used during training, and when $E(S_1)$ and $E(S_2)$ are
similar, their composition $\psi^{2} \circ (\psi^{1})^{-1}$ aligns $E(S_1)$ to
$E(S_2)$ and produces accurate correspondences (\cref{fig:before_after_flow}).
For this reason, points with similar embeddings from different shapes are mapped
to nearby regions within the anchor (green box in \cref{fig:teaser}), provided
that both shapes share the same semantic structure.

% GIULIOOOOOO
% This same reasoning delimits when \repname is most useful. When $\mathcal{S}_1$
% and $\mathcal{S}_2$ differ by a transformation to which $E$ is not invariant,
% $E(\mathcal{S}_1)$ and $E(\mathcal{S}_2)$ become dissimilar, degrading
% embedding-based methods such as KNN; by aligning their images through the anchor
% before matching, \repname reduces this sensitivity. Hence, as isometry decreases
% (\cref{tab:mesh-mesh-results}) or landmark placement becomes noisy
% (\cref{fig:stability_combined}), the advantage of \repname over other methods
% tends to increase.

% LoreNAM
The same reasoning also highlights when~\repname can be effective. When
$\mathcal{S}_1$ and $\mathcal{S}_2$ differ by a transformation to which $E$ is
not invariant, the distributions $E(\mathcal{S}_1)$ and $E(\mathcal{S}_2)$
become dissimilar and not alignable by the flow composition. Instead, when the
two embedding distributions share the same semantic structure, by forcing the
image of $E(\mathcal{S}_1)$ to align with $E(\mathcal{S}_2)$ before applying KNN
(\cref{fig:before_after_flow}),~\repname reduces the sensitivity to embedding
variations. This is why, as the isometry decreases
(\cref{tab:mesh-mesh-results}) or the landmark placement becomes noisy
(\cref{fig:stability_combined}), the advantage of~\repname over other methods
tends to increase.

\begin{figure}[!t]
\centering

\begin{tikzpicture}

% ---- First Plot: Delta Perturbation ----
\begin{axis}[
    name=plot1,
    width=0.54\textwidth,
    height=4cm,
    xlabel={$\delta$},
    ylabel={Error (\%)},
    xmin=0.0, xmax=0.2,
    ymin=0.01, ymax=0.1,
    xtick={0.0,0.05,0.1,0.15,0.2},
    ytick={0.01, 0.04, 0.07, 0.1},
    xticklabels={0, 5, 10, 15, 20},
    yticklabels={1, 4, 7, 10},
    scaled y ticks=false,
    ymajorgrids=true,
    grid style=dashed,
]

\addplot[color=red, mark=square, smooth] coordinates {
    (0.0,0.0338)
    (0.01,0.0369)
    (0.02,0.0374)
    (0.05,0.0546)
    (0.1,0.0627)
    (0.2,0.0838)
};

\addplot[color=blue, mark=o, smooth] coordinates {
    (0.0,0.0236)
    (0.01,0.0256)
    (0.02,0.0266)
    (0.05,0.0341)
    (0.1,0.0346)
    (0.2,0.0546)
};

\end{axis}

% ---- Second Plot: Landmark Number ----
\begin{axis}[
    at={(plot1.right of south east)},
    anchor=left of south west,
    xshift=0cm,
    width=0.54\textwidth,
    height=4cm,
    xlabel={Number of Landmarks},
    xmin=3, xmax=30,
    ymin=0.02, ymax=0.055,
    xtick={5,10,15,20,25,30},
    ytick={0.02, 0.03, 0.04, 0.05},
    yticklabels={2, 3, 4, 5},
    scaled y ticks=false,
    legend pos=north east,
    ymajorgrids=true,
    grid style=dashed,
    legend style={draw=none, fill=white}
]

\addplot[color=red, mark=square, smooth] coordinates {
    (3,0.051)
    (5,0.044)
    (7,0.034)
    (9,0.030)
    (11,0.028)
    (13,0.027)
    (15,0.026)
    (17,0.026)
    (20,0.026)
    (25,0.0256)
    (30,0.0247)
};
\addlegendentry{KNN}

\addplot[color=blue, mark=o, smooth] coordinates {
    (3,0.046)
    (5,0.032)
    (7,0.0264)
    (9,0.0267)
    (11,0.026)
    (13,0.0245)
    (15,0.0239)
    (17,0.0229)
    (20,0.0219)
    (25,0.0230)
    (30,0.0212)
};
\addlegendentry{\repname}

\end{axis}

\end{tikzpicture}

\caption{Stability analysis of FUSE.
On the left: robustness under landmark perturbation ($\delta$). On the right: performances varying the number of landmarks.}
\label{fig:stability_combined}

\end{figure}

\subsection{Embedding Selection}\label{sec:embedding_sel}

\begin{wraptable}[12]{r}{.4\linewidth}
   
\centering

\vspace{-1.1cm}

\caption{
    Correspondence error under different prior assumptions. The results are
    obtained without any test-time refinement. *ULRSSM results are obtained
    using public checkpoints. Details on metric and datasets in
    \ref{sec:evaluation}.
}
\begin{tikzpicture}
    
\node[scale=0.70] {
\begin{tabular}{ll|cc}
\toprule
& & \multicolumn{2}{c}{Error} \\
Method & Prior & FAUST & SMAL \\
%\cmidrule(l){1-2} \cmidrule(l){3-4} 
\midrule
Diff3 & Vision model & 0.089 & 0.189 \\
ULRSSM* & Training-data & 0.043 & 0.063 \\
\repname & Landmarks & \textbf{0.028} & \textbf{0.059} \\
\bottomrule
\end{tabular}
\label{tab:prior_comparison}
};

\end{tikzpicture}

\end{wraptable}

The choice of embedding $E$ is critical. First, the generalization of \repname
to any 3D representation depends on $E$ being computable and invariant over 3D
representations. Secondly, the embedding determines the properties of the
\repname map (\textit{e.g.}, shape matching needs some $E$ that captures
metric and semantic information about the shapes). Every shape matching pipeline
relies on assumptions that inject prior knowledge into the process. These
assumptions are necessary to resolve fundamental ambiguities, most notably
intrinsic symmetries, which cannot be disambiguated from purely geometric
information alone. Past methods encode this prior in different ways:
\cite{cao23unsup} learns a feature extractor in an unsupervised manner, where
the (data) prior is induced by training over a collection of shapes;
\cite{Dutt_2024_CVPR} leverages a pretrained large-scale vision model to
extract semantic information from rendered views of meshes;
\cite{aubry2011wave} relies on informative spectral signatures. Each of these
strategies introduces representation-specific dependencies that limit their
applicability: spectral bases, rendering pipelines, or available training data.

% In multi-representation settings, such priors are not directly applicable, making it challenging to design universally computable features.
% For triangle meshes, several learned encodings exist leveraging spectral signatures~\cite{sun2009concise, aubry2011wave}, neural architectures~\cite{diffusionnet}, and correspondence supervision~\cite{donati2022deep, cao23unsup}. These embeddings are effective for matching but rely on geometric priors which are precise only on meshes (e.g., Laplacian eigenfunctions) or supervised training.
% However, in multi-representation scenarios, usual geometric or data priors are
% not readily exploitable. In these cases, designing or learning universally
% computable features becomes non-trivial. 
In contrast, 3D coordinates (\ie, ``identity'' embeddings) are universally
available but lack intrinsic geometric or semantic meaning, making them
unsuitable for shape correspondence. One effective alternative is the Geodesic
Feature Embedding (\textbf{GFE}), which represents surface points as
vectors of geodesic distances to a small set of landmarks $L$, yielding a compact
embedding in $\mathbb{R}^{|L|}_{+}$ that encodes the intrinsic geometry.
Landmarks provide a minimal yet powerful form of prior: they provably resolve
intrinsic symmetries by anchoring the semantic orientation of the shape, and are
consistently employed in axiomatic shape matching methods derived from the
Functional Maps~\cite{ovsjanikov2012functional} framework. In our case,
landmarks are manually selected on one shape (5 or 6 depending on the semantics
of the dataset) and transferred to the rest of the dataset via ground truth
correspondences; \cref{fig:landmarks} in the supplementary material shows the
selected landmarks for each dataset. As shown in
\cref{tab:prior_comparison}, despite this simple prior, \repname{} with
landmark-based GFE achieves the lowest correspondence error on both FAUST and
SMAL against state-of-the-art prior-based methods~\cite{Dutt_2024_CVPR,
cao23unsup}.

Importantly, different algorithms have been developed to compute or efficiently
approximate geodesic distances on triangle meshes, signed distance fields, and
point clouds, such as Dijkstra's shortest path algorithm, the heat
method~\cite{heatmethod}, and fast marching~\cite{MEMOLI2001730}. While
computing geodesic distances generally involves estimating the underlying
surface, established methods exist (e.g.~\cite{heatmethod}) that do so
implicitly with stable results. As a consequence, \repname does not require
explicit surface estimation over point clouds and can employ them directly as an
input representation.

However, geodesic embeddings are not naturally aligned across shapes from a
distributional perspective as a non-rigid deformation can significantly alter
the geodesic distances between points. With \repname, the flow composition acts
as a probabilistic alignment mechanism: by mapping geodesic features through the
Gaussian prior, we obtain distributional alignment across shapes, as shown in
\cref{fig:before_after_flow} \cref{fig:before_after_flow}.

% \begin{table}[]
% \centering
% \caption{}
% \label{tab:embedding_comparison}
% \begin{tabular}{@{}lcccc@{}}
% \toprule
% \multirow{2}{*}{\textbf{Embedding}} & \multicolumn{2}{c}{\textbf{KNN}}                      & \multicolumn{2}{c}{\textbf{\repname}}                 \\ \cmidrule(l){2-5} 
%                                     & \textbf{Acc. $\downarrow$} & \textbf{Sim. $\uparrow$} & \textbf{Acc. $\downarrow$} & \textbf{Sim. $\uparrow$} \\ \midrule
% xyz (3)                             & 0.0203                     & 0.8410                   & 0.0308                     & 0.9131                   \\
% wks (10)                            & 0.0186                     & 0.9048                   & 0.0130                     & 0.9711                   \\
% geod (5)                            & 0.0034                     & 0.9436                   & 0.0021                     & 0.9870                   \\
% FMNet (30)                          & 0.0001                     & 0.9902                   & 0.0012                     & 0.9887                   \\ \bottomrule
% \end{tabular}
% \end{table}

% \begin{table}[htbp]
% \centering
% \caption{Performance comparison for FAUST and SMAL datasets}
% \label{tab:my-table}

\begin{table}[!t]
\centering
\caption{Euclidean error on FAUST and SMAL datasets with different embeddings. WKS* and WKS-L* are smoothed versions, to show the improvement of FUSE over KNN in the presence of low-energy embeddings.}
\label{tab:embedding-comparison}
\renewcommand{\arraystretch}{1.2}
\resizebox{\textwidth}{!}{%
\begin{tabular}{l cccccccccccc}%{l *{7}{S[table-format=1.4]} *{5}{S[table-format=1.4]}}
\toprule
& \multicolumn{7}{c}{\textbf{FAUST}} & \multicolumn{5}{c}{\textbf{SMAL}} \\
\cmidrule(lr){2-8} \cmidrule(lr){9-13}
\textbf{Method}
  & {\textbf{$x, y, z$}} & {\textbf{GFE}} & {\textbf{WKS}} & {\textbf{WKS*}} & {\textbf{WKS-L}} &  {\textbf{WKS-L*}} & {\textbf{ULRSSM}}
  & {\textbf{$x, y, z$}} & {\textbf{GFE }} & {\textbf{WKS}} & {\textbf{WKS-L}} & {\textbf{ULRSSM}} \\
\midrule
Dim
  & 3 &5 & 20 & 20 & 20 & 20 & 256
  & 3 & 5 & 20 & 20 &  256 \\
\midrule
KNN
  & 0.1572 & 0.0454 & 0.1007 & 0.1125 & 0.0428 & 0.0522 & 0.0723
  & 0.2903 & 0.0879 & 0.2736 & 0.0697 &   0.1061 \\
\repname
  & 0.1488 & \textbf{0.0289} & 0.1015 & 0.0901  & 0.0556 & 0.0416 & 0.0829
  & 0.3226 & \textbf{0.0595} & 0.3039 & 0.0668 & 0.1117  \\
\midrule
Dirichlet
  & 0.0002  & 0.0001 & 0.0056  & 0.0008 & 0.0022  & 0.0004 &   0.1847
  &  0.0001 & 0.0001  & 0.0257 & 0.0131 & 0.637  \\
\bottomrule
\end{tabular}}
\end{table}

% \begin{table}[!t]
% \centering
% \caption{
%     Comparison of embedding strategies (dimensionality reported aside) in
%     terms of correspondence accuracy ($\downarrow$) and distributions similarity ($\uparrow$). Results computed over 45 isometric pairs.
% }
% \label{tab:embedding_comparison}
% \footnotesize
% % \begin{tikzpicture}
% % \node[scale=0.9] {
%     \begin{tabular}{lcccc}
%         \toprule
%         \multirow{2}{*}{\textbf{Embedding}} 
%          & \multicolumn{2}{c}{\textbf{KNN}} 
%          & \multicolumn{2}{c}{\textbf{\repname}} \\
%          & \textbf{Acc. $\downarrow$} & \textbf{Sim. $\uparrow$} & \textbf{Acc. $\downarrow$} & \textbf{Sim. $\uparrow$} \\
%         \cmidrule(lr){1-1} \cmidrule(lr){2-3} \cmidrule(lr){4-5}
%         xyz   (3)          & 0.020316 & 0.841044 & 0.030804 & 0.913107 \\
%         wks   (10)          & 0.018615 & 0.904846 & 0.013043 & 0.971122 \\
%         geod  (5)          & 0.003442 & 0.943582 & 0.002094 & 0.986980 \\
%         FMNet (30)  & 0.000062 & 0.990156 & 0.001158 & 0.988705 \\
%         \bottomrule
%     \end{tabular}
% % };
% % \end{tikzpicture}
% \end{table}
\input{Figures/table+rendering}

\paragraph{Generality and robustness.}
While GFE is our primary choice, \repname can be applied to any pointwise
embedding. To understand when it is effective, we vary both the embedding type
(\cref{tab:embedding-comparison}) and the landmark configuration
(\cref{fig:stability_combined}). In our experiments,~\repname does not appear to
suffer from the increasing dimensionality: adding landmarks from 3 to 30 yields
progressively better results that consistently outperform KNN
(\cref{fig:stability_combined}, right); we observe the same behaviour when
perturbing landmark positions by a random fraction of the shape diameter
(\cref{fig:stability_combined}, left). Instead, \cref{tab:embedding-comparison}
reveals that a critical factor is \emph{smoothness}: GFE has very low Dirichlet
energy and benefit greatly from the flow mapping, while WKS~\cite{aubry2011wave}
and learned embeddings~\cite{cao23unsup} have substantially higher Dirichlet
energy, resulting in poor performance. We refer to this as \emph{curse of
frequency}: high-frequency embeddings induce complex distributions that are
harder for the flow to align, a challenge mirroring known difficulties in
Functional Maps optimization~\cite{melzi2019zoomout, ovsjanikov2012functional}.
GFE exhibits lower spectral complexity, making them particularly well-suited for
FUSE. We note that smoothing the embeddings via projection on a small set of
Laplace-Beltrami eigenbasis~\cite{Taubin,Levy06,Levy08}, (WKS* and WKS-L* in
\cref{tab:embedding-comparison}) consistently improves FUSE's accuracy over raw
features, even in non-landmark-based scenarios. This suggests that integrating
smoothly learned features with flow-based models is a viable direction and
fosters future work. 

%\paragraph{Bijectivity.}
%The invertibility of the flows does not guarantee a bijection between the
%shapes. We use pointwise embeddings to maintain a one-to-one relationship
%between points on the shape and their representation in the embedding space.
%Since embeddings are treated as vertex attributes, invertibility of the
%embedding computation is not required, nor is it affected by its smoothness.
%After aligning the source and target embeddings via flow composition, we apply
%KNN to each source embedding to retrieve the nearest target point as a match
%(\cref{eq:knn}). Due to this KNN step, the overall matching is not strictly
%bijective, as is standard in point-based correspondence methods.

We note that the invertibility of the flows does not by itself induce a
bijection between the shapes: the final KNN step (\cref{eq:knn}) assigns each
source point its nearest target independently, so the matching is not strictly
bijective, as is standard in point-based correspondence. Since both the GFE and
the flow are defined continuously over the surface, the construction naturally
extends toward a continuous, fully bijective map, which we leave to future work.

%However, since
%both the GFE and the flow are defined continuously over the surface, our
%construction already achieves strong results and theoretically supports a
%continuous solution. 

\subsection{Flow matching over Diffusion}
\label{sec:GD}
Building on the recent work of~\cite{zhang2024geometry}, which represents 3D
shapes as probability distributions generated via a Denoising Diffusion Implicit
Model (DDIM)~\cite{song2022denoisingdiffusionimplicitmodels}, we adopt the same
probabilistic representation while constructing maps between shapes using flow
matching rather than DDIM. Although FM and DDIM can be similar, the former
offers distinct advantages for our purposes. FM establishes an explicit
relationship between the Gaussian distribution and the data distribution. This
ensures that the inverse flow maps any shape back to the exact Gaussian
distribution in a deterministic manner. This property guarantees that all shapes
lie in an ideally shared latent space. In contrast, methods such as DDIM only
asymptotically approach the Gaussian through large noise schedules, resulting in
deviations from the anchor space.

\begin{wraptable}[11]{r}{.45\textwidth}
\vspace{-1.4cm}
\centering

\caption{
    \repname\textsubscript{DDIM} vs \repname in terms of shape matching error (first four rows) and probability divergence (last two rows) on the FAUST dataset. Lower is better for all the metrics.
}\label{tab:comparison}

\begin{tikzpicture}
\node[scale=0.65] {
\begin{tabular}{l c c c}
    \toprule
    \textbf{Metric } & 
    \textbf{Raw} & 
    \textbf{\repname\textsubscript{DDIM}} & 
    \textbf{\repname} \\
    \cmidrule(lr){1-1} \cmidrule(lr){2-4} 
    Euclidean Error (anchor) & 0.045 & 0.099 & 0.042 \\
    Geodesic Error (anchor) & 0.043 & 0.095 & 0.039 \\

    Euclidean Error & - & 0.037 & 0.028 \\
    Geodesic Error & - & 0.035 & 0.027 \\
    
    \cmidrule(lr){1-1} \cmidrule(lr){2-4} 
    Mean KL Divergence & 0.2528 & 0.3519 & 0.0375 \\
    Mean JS Divergence & 0.1928 & 0.2128 & 0.0829 \\
    %Mean Wasserstein Distance  & 0.0201 & 0.3744 & 0.0482 \\
    %MMD Distance  & 0.0075 & 0.1321 & 0.0334 \\
    %Energy Distance   & 0.0335 & 0.5872 & 0.0540 \\
    \bottomrule
\end{tabular}
};
\end{tikzpicture}

\end{wraptable}
We quantify this effect in \cref{tab:comparison}, evaluating shape pairs with GFE using two criteria: (1)
the accuracy of nearest-neighbor correspondences both in the anchor space and in
the embedding space, and (2) the JS and KL divergence between embedding
distributions. We compute these metrics before and after inverting the flows
using DDIM (\repname~\textsubscript{DDIM}) and FM (\repname). The results show
that \repname achieves substantially lower divergence and higher correspondence
accuracy, confirming that accurate anchoring to the Gaussian directly correlates
with improved matching performance. DDIM's inversion leads to larger deviations
from the anchor and thus inferior map quality. In the supplementary materials,
we provide a mathematical derivation of this crucial difference, along with
additional examples.

\section{Results}\label{sec:evaluation} We evaluate our
representation, considering shape matching as the main task, but also presenting
possible extensions of \repname targeting other applications.

%In this section, we evaluate our approach in the context of non-rigid shape
%matching, considering standard benchmarks a a o scenario hetchingn, we apply
%\repname as a backbone providing novel solutions for multiple
%inter-representation applications.

\input{Figures/matching_qualitative}

\subsection{Non-Rigid Shape Matching}\label{sec:mesh-shape-matching}
\input{Figures/multi_repr_match}
We first analyze the performance of \repname against more standardized
benchmarks, considering pairs of meshes and point clouds. We then explore novel
and more challenging scenarios by matching SDFs and inter-representation pairs.
We restrict our analysis to landmarks-based methods, as discussed in
\cref{sec:embedding_sel}.
%This section presents the quantitative evaluation of our method in the context
%of pairwise shape matching. We first compare our approach in a standard setting,
%where both source and target shapes come from the same representation, i.e.,
%meshes for FAUST, SMAL, and SHREC20 while point clouds for KINECT. Then, we explore the
%novel capabilities of \repname by evaluating its performance in the context of
%SDFs and inter-representation matching.

\paragraph{Baselines.}
%\lorenzo{I need some help to motivate the choise of the baselines}\\
%To quantitatively evaluate our method in the context of non-rigid shapa
%matching, we select a range of methods from the literature that can be used to
%match features. Our baselines include relatively naive approaches like KNN and
%OT with Sinkhorn
%relaxation~\cite{cuturi2013sinkhorndistanceslightspeedcomputation}, functional
%maps with WKS descriptors~\cite{ovsjanikov2012functional} and with geodesic
%distance descriptors, learning-based refinement methods like the Zoomout
%algorithm (ZO)~\cite{melzi2019zoomout} and NAM~\cite{vigano2025nam}, and the
%fully extrinsic Neural Deformation Pyramid (NDP)~\cite{li2022nonrigid} optimized
%with each shape's input landmarks. Finally, we compare two variants of our
%method obtained by further refining the point-to-point map obtained with
%\repname using ZO and NAM.
We evaluate~\repname against baselines from different paradigms. Methods using
our same geodesic feature encoding include nearest neighbor search in the
embedding space (KNN) and in the anchor distribution space
(\repname~\textsubscript{ANCHOR}), OT (Optimal Transport with Sinkhorn
relaxation~\cite{cuturi2013sinkhorndistanceslightspeedcomputation}), Functional
Maps (FMaps) with geodesic descriptors~\cite{ovsjanikov2012functional}, and
refined variants FMaps~\textsubscript{ZO} (ZoomOut\cite{melzi2019zoomout}) and
FMaps~\textsubscript{NAM} (Neural Adjoint Maps~\cite{vigano2025nam}). Further, we
also consider the original Functional Maps formulation with
WKS~\cite{aubry2011wave}, the neural extrinsic Neural Deformation Pyramid (NDP)
approach~\cite{li2022nonrigid}, the vision-foundation
Diff3F~\cite{Dutt_2024_CVPR}, and the data-driven ULRSSM~\cite{cao23unsup}.
Finally, we use \repname to initialize ZoomOut and NAM
(\repname~\textsubscript{ZO} and \repname~\textsubscript{NAM}).

\paragraph{Datasets.}
We evaluate our method on well-established datasets for non-rigid shape
matching:  
\textbf{FAUST}~\cite{bogo2014faust} (near-isometric human meshes),
\textbf{SMAL}~\cite{Zuffi2017,cao23unsup} (non-isometric quadruped meshes),
\textbf{SHREC20}~\cite{shrec20} (strongly non-isometric animal meshes), and
\textbf{KINECT}~\cite{bhatnagar2022behavedatasetmethodtracking,vigano2025nam}
(human point clouds). We manually pick 5 landmarks for humans and 6 for animals,
\cref{fig:landmarks} in the supplementary shows the selected landmarks for each
dataset.

\paragraph{Metrics.}
We evaluate the quality of correspondence using four metrics: two measures of
accuracy, Euclidean (Eucl.) and geodesic (Geod.) distances~\cite{kim2011blended},
as well as Dirichlet energy~\cite{magnet2022smooth} and coverage of the computed
maps. Lower values are better for the first three metrics, while a higher value
is better for Coverage. Quantitative results are reported in
\cref{tab:mesh-mesh-results}, qualitative in \cref{fig:qual_res}, and plots with
the full distribution of the metrics in~\cref{fig:boxplots} of the supplementary
material.

\begin{table}[t]
\centering
\caption{
    Averaged results from all the considered datasets for our method and the
    other baseline solutions. For each column, the best result is highlighted in
    bold.
}
\label{tab:mesh-mesh-results}
\resizebox{\columnwidth}{!}{%
\begin{tabular}{@{}lcccc|cccc|cccc|cccc@{}}
\toprule
 & \multicolumn{4}{c}{FAUST (quasi-iso)} 
 & \multicolumn{4}{c}{SMAL (non-iso)} 
 & \multicolumn{4}{c}{SHREC20 (strong non-iso)} 
 & \multicolumn{4}{c}{Kinect (point cloud)} \\ 
\cmidrule(l){2-17} 
\multirow{-2}{*}{Method} 
& Eucl. $\downarrow$ & Geod. $\downarrow$ & Dirich. $\downarrow$ & \multicolumn{1}{c|}{Cov. $\uparrow$} 
& Eucl. $\downarrow$ & Geod. $\downarrow$ & Dirich. $\downarrow$ & \multicolumn{1}{c|}{Cov. $\uparrow$} 
& Eucl. $\downarrow$ & Geod. $\downarrow$ & Dirich. $\downarrow$ & \multicolumn{1}{c|}{Cov. $\uparrow$} 
& Eucl. $\downarrow$ & Geod. $\downarrow$ & Dirich. $\downarrow$ & Cov. $\uparrow$ \\ 
\midrule
KNN
& 0.0454 & 0.0434 & 0.0022 & \multicolumn{1}{c|}{0.2452}
& 0.0879 & 0.0836 & 0.0016 & \multicolumn{1}{c|}{0.1364}
& 0.1083 & 0.1036 & 0.0063 & \multicolumn{1}{c|}{0.1446}
& 0.1119 & 0.1095 & 0.0025 & 0.0995 \\

OT~\cite{cuturi2013sinkhorndistanceslightspeedcomputation}
& 0.0926 & 0.0856 & 0.0016 & \multicolumn{1}{c|}{0.0890}
& 0.0779 & 0.0741 & 0.0012 & \multicolumn{1}{c|}{0.1129}
& 0.1252 & 0.1168 & 0.0060 & \multicolumn{1}{c|}{0.1001}
& 0.1327 & 0.1273 & 0.0047 & 0.0748 \\

FMap~\cite{ovsjanikov2012functional}
& 0.0401 & 0.0400 & 0.0016 & \multicolumn{1}{c|}{0.3236}
& 0.0756 & 0.0744 & 0.0018 & \multicolumn{1}{c|}{0.2162}
& 0.1381 & 0.1411 & 0.0077 & \multicolumn{1}{c|}{0.2037}
& 0.0810 & 0.0779 & 0.0031 & 0.1683 \\

FMap\textsubscript{WKS}~\cite{aubry2011wave}
& 0.0331 & 0.0334 & 0.0017 & \multicolumn{1}{c|}{0.3759}
& 0.0647 & 0.0623 & 0.0023 & \multicolumn{1}{c|}{0.2585}
& 0.1197 & 0.1131 & 0.0203 & \multicolumn{1}{c|}{0.2380}
& 0.0723 & 0.0718 & 0.0045 & 0.2211 \\

FMap\textsubscript{ZO}~\cite{melzi2019zoomout}
& 0.0207 & 0.0195 & \textbf{0.0009} & \multicolumn{1}{c|}{0.6969}
& 0.0625 & 0.0611 & 0.0065 & \multicolumn{1}{c|}{0.4706}
& 0.1454 & 0.1480 & 0.0152 & \multicolumn{1}{c|}{0.3495}
& 0.0706 & 0.0710 & 0.0062 & 0.4043 \\

FMap\textsubscript{NAM}~\cite{vigano2025nam}
& 0.0245 & 0.0240 & \textbf{0.0009} & \multicolumn{1}{c|}{0.6323}
& 0.0617 & 0.0589 & 0.0015 & \multicolumn{1}{c|}{0.4694}
& 0.1041 & 0.0962 & 0.0082 & \multicolumn{1}{c|}{0.2860}
& 0.0689 & 0.1118 & 0.0026 & 0.3494 \\

NDP~\cite{li2022nonrigid}
& 0.0702 & 0.0886 & 0.0020 & \multicolumn{1}{c|}{0.4893}
& 0.0555 & 0.0675 & \textbf{0.0010} & \multicolumn{1}{c|}{0.4818}
& 0.0698 & 0.0649 & \textbf{0.0013} & \multicolumn{1}{c|}{0.5016}
& \textbf{0.0457} & \textbf{0.0438} & \textbf{0.0012} & 0.4190 \\

Diff3f~\cite{Dutt_2024_CVPR}
& 0.0697 & 0.1109 & 0.0497 & \multicolumn{1}{c|}{0.3075}
& 0.1594 & 0.2278 & 0.0560 & \multicolumn{1}{c|}{0.2146}
& 0.0876 & 0.1036 & 0.0443 & \multicolumn{1}{c|}{0.1579}
& 0.1394 & 0.1732 & 0.0755 & 0.1448 \\

Diff3f FMap~\cite{Dutt_2024_CVPR}
& 0.0553 & 0.0886 & 0.0036 & \multicolumn{1}{c|}{0.4427}
& 0.1283 & 0.1943 & 0.0114 & \multicolumn{1}{c|}{0.2789}
& 0.1173 & 0.1357 & 0.0133 & \multicolumn{1}{c|}{0.2376}
& 0.1880 & 0.2087 & 0.0362 & 0.0565 \\

ULRSSM~\cite{cao23unsup}
& \textbf{0.0092} & \textbf{0.0109} & 0.0011 & \multicolumn{1}{c|}{\textbf{0.7302}}
& \textbf{0.0370} & \textbf{0.0374} & 0.0030 & \multicolumn{1}{c|}{0.4936}
& 0.3286 & 0.3496 & 0.2299 & \multicolumn{1}{c|}{0.1337}
& N/A & N/A & N/A & N/A \\

\midrule

% \textsc{knn-in-gauss}
FUSE\textsubscript{ANCHOR}
& 0.0425 & 0.0390 & 0.0241 & \multicolumn{1}{c|}{0.3483}
& 0.0579 & 0.0542 & 0.0185 & \multicolumn{1}{c|}{0.3708}
& 0.0912 & 0.0806 & 0.0538 & \multicolumn{1}{c|}{0.3619}
& 0.0983 & 0.0925 & 0.0328 & 0.2459 \\

% \repname-{DDIM}
\repname\textsubscript{DDIM}
& 0.0402 & 0.0407 & 0.0023 & \multicolumn{1}{c|}{0.2973}
& 0.0921 & 0.0879 & 0.0016 & \multicolumn{1}{c|}{0.1500}
& 0.0856 & 0.0797 & 0.0030 & \multicolumn{1}{c|}{0.2214}
& 0.1106 & 0.1101 & 0.0025 & 0.1326 \\

\textbf{\repname}
& 0.0289 & 0.0274 & 0.0066 & \multicolumn{1}{c|}{0.5320}
& 0.0595 & 0.0576 & 0.0075 & \multicolumn{1}{c|}{0.5156}
& 0.0975 & 0.0896 & 0.0339 & \multicolumn{1}{c|}{0.5278}
& 0.0969 & 0.0914 & 0.0137 & \textbf{0.4931} \\

% \textbf{\repname-{ZO}}
\textbf{\repname\textsubscript{ZO}}
& 0.0200 & 0.0189 & \textbf{0.0009} & \multicolumn{1}{c|}{0.7127}
& 0.0437 & 0.0413 & 0.0023 & \multicolumn{1}{c|}{0.5361}
& 0.1275 & 0.1312 & 0.0093 & \multicolumn{1}{c|}{0.3763}
& 0.0690 & 0.0662 & 0.0046 & 0.4173 \\

\textbf{\repname\textsubscript{NAM}}
& 0.0179 & 0.0168 & \textbf{0.0009} & 0.7027
& 0.0422 & 0.0388 & \textbf{0.0010} & \textbf{0.5520}
& \textbf{0.0658} & \textbf{0.0603} & 0.0027 & \textbf{0.5342}
& 0.0770 & 0.0702 & 0.0023 & 0.3904 \\

\bottomrule
\end{tabular}
}
\end{table}

\paragraph{Mesh correspondences.} 
While FMaps-based methods tend to produce smoother maps with smaller Dirichlet
energy, our method standouts in terms of both Euclidean and geodesics accuracy.
Due to the nature of the flows, invertible by design, \repname achieves better
coverage of the target surface overall. The introduction of refinement
strategies (\ie, ZO and NAM) adds pairwise optimization, further improves the
base results of the~\repname pipeline and while considerably increasing the
smoothness of the maps. Notably, initializing these refinements methods with
\repname yields better results than using the standard FMaps initialization,
particularly evident in the case SHREC20 with strongly non-isometric pairs.
Geodesic distances are particularly sensitive to deviations from isometry, and
even though corresponding points can be associated with structurally similar
distance distributions, actual values may wildly vary. We also observe that, in
the more challenging scenario of non-isometric pairs,
\repname~\textsubscript{ANCHOR} achieves marginally better accuracy than our
approach, at the cost of a considerably higher Dirichlet energy and lower
coverage. This suggests that the lack of smoothness in the maps produced by
\repname depends on the flow inversion from the surface to the anchor
distribution, which is a non-smooth operation.
ULRSSM~\cite{cao23unsup} achieves the best results on FAUST and SMAL, but
under performs substantially on SHREC20. 
As a data-driven method, it requires extensive training on a large dataset, it is restricted to meshes, whereas our flows can be efficiently trained from different representations and easily composed.

\paragraph{Point cloud correspondences.}
%FUSE requires reliable geodesic distances and sufficient surface coverage, both of which are difficult to guarantee on unstructured data. Geodesics must be approximated over k-nearest-neighbor graphs, introducing noise that propagates into the embeddings and the learned flows. This degrades FUSE's performance relative to extrinsic methods like NDP [21], which bypasses geodesics entirely — though at the cost of failing under strong non-rigid deformations (e.g., FAUST). Nonetheless, FUSE consistently outperforms non-refined geodesic baselines (KNN, OT, FMaps) on point clouds, confirming that the flow alignment adds value even with noisy embeddings. Improving geodesic estimation on point clouds, for instance through learned intrinsic features, is a natural direction for future work.

Point cloud matching presents a unique challenge: FUSE requires reliable
geodesic distances and sufficient surface coverage, both of which are difficult
to guarantee on unstructured data, since the surface information outside of the
vertices is unknown. This degrades FUSE's performance relative to extrinsic
methods like NDP~\cite{li2022nonrigid}, which bypasses geodesics entirely at the
cost of failing under strong non-rigid deformations (e.g., FAUST). However,
\repname still outperforms non-refined geodesic methods (KNN, OT, FMaps),
proving that the flow mechanism itself adds value. 
% This limitation reveals the broader challenge: computing intrinsic quantities on point clouds remains an open problem. Moreover we note that in the case of point clouds, we do not have access to the surface information, and this returns to a non uniformity of the points sampled during training of the flows.

% \loreNAM{
% \textbf{If we don't I can surely write it better, we might add this to the SDFs section if want to evaluate the matching on meshes}
% Although the matching process is perfomed in the intrinsic space of the
% geodesic features, the quality of the correspondences between the
% vertices of the meshes is evaluated in the original 3D space.

% Because of the bijective nature of this mapping, we find the permutation solving
% the matching problem in the features space, then, we apply this permutation to
% the corresponding mesh vertices for evaluation.

% This is particularly important when we consider matching between
% \emph{non-meshes}. We use features computed from the points in the respective
% domains, for example, in the case of SDFs, we sample points on the zero-level
% set of the SDF to solve the correspondences between the shapes. Then, once a
% permutation between the shapes is found, we apply it to the vertices of the
% corresponding mesh for evaluation.
% }

\subsection{Alternative Representations}\label{sec:alternative-representations}
Since \repname only requires the ability to sample points from a surface and to
compute single-to-all geodesic distances, it enables shape matching in a broader
class of representations, such as Signed Distance Functions (SDFs), as well as
inter-representation matching (see \cref{fig:multi_repr_match}). 
% Figure~\ref{fig:multi_repr_match} provides a qualitative overview of these novel inter-representation matching scenarios.

\paragraph{Inter-representation matching.}

We conduct experiments on shape pairs from different FAUST-like collections:
from the original FAUST meshes, we extract the vertices to use as point clouds,
and we compute SDFs as discussed in the previous paragraph. 
For each shape,
geodesics are computed sticking to the constraints of the representation (\ie,
heat geodesics~\cite{heatmethod} on point clouds, Dijkstra on mesh or voxel
grid). As mentioned in \cref{sec:mesh-shape-matching}, \repname inevitably
struggles on matching tasks involving only point clouds, but achieves results
comparable to mesh-to-mesh matching when at least one shape is represented as
mesh or SDF. We provide an example of matching for all possible combinations of
meshes, point clouds and SDFs in \cref{fig:multi_repr_match}. A deeper analysis
is presented in the supplementary material.

\paragraph{SDFs shape matching.}
% \usepackage{booktabs}
% \usepackage{graphicx}
\begin{wraptable}[9]{l}{.4\textwidth}
\centering

\vspace{-1.28cm}

\caption{
    Quantitative evaluation of the SDF-based FAUST dataset without considering the topological error.
}\label{tab:sdf-sdf-eval}
% \resizebox{\columnwidth}{!}{%
\begin{tikzpicture}
\node[scale=0.64] {
\begin{tabular}{@{}lcccc@{}}
\toprule
Method       & Eucl. $\downarrow$ & Geod. $\downarrow$ & Dirich. $\downarrow$ & Cov. $\uparrow$ \\ \midrule
KNN          & 0.0562           & 0.0569          & 0.0023          & 0.1744          \\
NDP     & 0.0525           & 0.0643          & \textbf{0.0016} & \textbf{0.5139} \\
OT          & 0.0921           & 0.0881          & 0.0018          & 0.0936          \\ 
\repname~\textsubscript{ANCHOR} & 0.0472           & 0.0469          & 0.0175          & 0.2784          \\
\textbf{\repname}         & \textbf{0.0375}  & \textbf{0.0385} & 0.0069          & 0.4653          \\
\bottomrule
\end{tabular}%
};
\end{tikzpicture}
\end{wraptable}
To our knowledge, no dataset exists to evaluate quantitatively matching
between SDFs, so we introduce a new benchmark derived from FAUST. For each
shape, we train a neural implicit model following~\cite{implicitarap}, and for
evaluation, we project the original mesh vertices onto the
corresponding SDF surface, ensuring accuracy is always measured on the
original mesh geometry. We note that this implicit reconstruction may alter the
topology of some shapes.
To apply \repname, we uniformly sample $\sim\!10^5$ points from the zero level
set and voxelize each SDF surface to compute geodesic distances via Dijkstra.
As in \cref{sec:mesh-shape-matching}, these geodesic embeddings are then used to
train the flow models. \cref{tab:sdf-sdf-eval} reports the results for all
shapes whose SDFs preserve valid topology.

\repname provides a matching estimation without pair-specific supervision,
additional training, or fine-tuning and with results comparable or better than
other methods. While ZO and NAM refinement methods can be applied between meshes
and point clouds to greatly improve the matching quality, these methods cannot
be used on SDFs where FUSE added value is most significant, as it provides the
most accurate maps compared to other baselines.

%To our knowledge, no dataset exists for quantitative evaluation of SDFs matching. Thus, for our analysis, we introduce a new benchmark derived from FAUST.
% To evaluate our method in the context of matching between SDFs, we create a new benchmark derived from the FAUST dataset. 
%We train a neural implicit model following~\cite{implicitarap}, and uniformly sample a large number of points ($\sim\hspace{-0.25em}10^5$) from the zero-level set. We voxelize the surface, computing geodesic distances on the resulting voxel grid using the Dijkstra algorithm. Analogously to \cref{sec:evaluation:mesh-pc}, the geodesic embeddings are used to train the flow models for \repname. We note that, with this procedure, the resulting SDF may have altered topology, when compared to the original mesh.
% does not preserve topological structure. Consequently, shapes such as the $6$s and $7$s in the FAUST dataset will exhibit altered topology when represented as SDFs compared to their original mesh.

% To quantitatively assess the quality of the correspondences obtained using our method, 
%For the evaluation, we project the vertices of the original FAUST meshes onto their respective SDF surfaces and compute the point-to-point correspondences with \repname. \cref{tab:sdf-sdf-eval} summarizes the results across the shapes with no topological errors.

\subsection{Applications}

%\subsection{Matching Across Representations}
%Our framework seamlessly supports correspondences across different shape representations:
\repname allows to model 3D surfaces as probability distributions via surface sampling and to compute maps without explicit optimization, effectively making it a ``bridge'' between distinct representations. We exploit this peculiar property to use \repname as a backbone for various tasks other than pure shape matching.
% can be employed as a ``bridge'' between distinct 3D representations, as it allows us to model 3D surfaces as probability distributions simply by samplingsurface points, and to compute maps among them in a zero-shot fashion. In this section, we conduct experiments using \repname as a backbone for various applications: \daniele{Remove this list after subsections are done}
% \begin{itemize}
%     \item \textbf{Human scans fitting and skinning}: Having access to a parametric model of human shapes like SMPL-X~\cite{SMPL-X:2019}, we can match a pointcloud to the model template and enable deformation transfer and other crucial application for digital humans.
%     \item \textbf{Mesh parametrization:} The analytical $uv$ parametrization of a sphere is transferred on a (genus zero) triangle mesh for texturing purposes.
%     \item \textbf{Volumetric matching:} We can apply our pipeline to Volumetric matching simply computing geodesic on volumetric shapes, for example using tetrahedral shapes.
% \end{itemize}

\paragraph{Human scans fitting and skinning.}
\input{Figures/smpl-fitting}

Given the scan of a human body, represented as a 3D point cloud $\mathcal{S}$,
we use \repname to match it to a triangle mesh representing the neutral pose
$\mathcal{T}$ of a templated body model, obtained with
SMPL-X~\cite{SMPL-X:2019}. By exploiting the correspondence, we can
easily fit the body model's parameters to approximate $\mathcal{S}$ via
minimization of the MSE between corresponding points.
%(rather than computing more complex energies, like Chamfer distance).
This procedure, whose results we show in the left column
of~\cref{fig:smpl-fitting}, yields style and pose vectors $(\beta,\theta)$. If
we compute the vertex offsets from the the SMPL-X output for the
optimized $\theta$ to the neutral pose $\mathcal{T}$, we can apply them ``as-is'' to the corresponding points in $\mathcal{S}$ to obtain the point cloud in neutral pose. 
Finally, given a novel pose vector $\theta'$, we can iterate this process to obtain
the scan in the required pose. Two example applications of our pipeline are
depicted in~\cref{fig:smpl-fitting}.
% we can edit the pose of the original scan $\mathcal{X}$ by exploiting the correspondence with $\mathcal{T}$: first, using the SMPL-X output for the optimized $\theta$, we compute the vertex-wise offsets induced by skinning on $\mathcal{T}$. Then, we subtract these from the corresponding points of $\mathcal{X}$ to obtain the original scan in neutral pose. Finally, we repeat the first step for $\theta'$ and sum the resulting offsets to the neutral-posed scan. The re-skinning results and a summary of the entire pipeline are showed in~\cref{fig:smpl-fitting}. 

%Visual inspection reveals that our method consistently aligns and achieves high coverage, even in the absence of explicit geometric priors or mesh topology, underscoring the generality of our flow representation approach.
\paragraph{Mesh parametrization.}
% We use \repname to transfer the analytical parametrization of a sphere onto genus-zero shapes. 

\input{Figures/mesh_parametrization}
Given a genus-zero surface $\mathcal{S}$ and a reference sphere $\Phi$, we train
flows from a Gaussian to each shape using 3D coordinates as embeddings.
Composing these flows yields a bijection between $\mathcal{S}$ and $\Phi$,
allowing us to transfer the sphere's UV parametrization to the shape.
As \repname maps may lack smoothness (see~\cref{sec:mesh-shape-matching}),
transferred UVs are used to supervise an MLP that predicts continuous UV
coordinates, enforcing a smooth prior. Since the MLP operates in the embedding
space, the procedure applies to any representation supported by \repname. In
this setting, 3D coordinates are sufficient, as no semantic consistency is
required. In \cref{fig:parametrization_pipeline} we show an example.

\paragraph{Volume matching.}\label{sec:volumetric}
Our representation-agnostic formulation naturally extends to volumetric shapes,
three-dimensional manifolds with well-defined geodesic distances, 
enabling targeting of volume-matching tasks. We apply \repname to tetrahedral
meshes, comparing it to Volumetric FMaps~\cite{maggioli2025volumetricfunctionalmaps}. Results from \cref{table:volumetric} and \cref{fig:volumetric} mirror our analysis on the surface scenario:
comparable or improved accuracy, particularly in case of strong
non-isometry, where the functional approach struggles.
Details on the metrics are in the supplementary materials.

\begin{minipage}{.54\textwidth}
\begin{figure}[H]
\centering
\includegraphics[width=\linewidth]{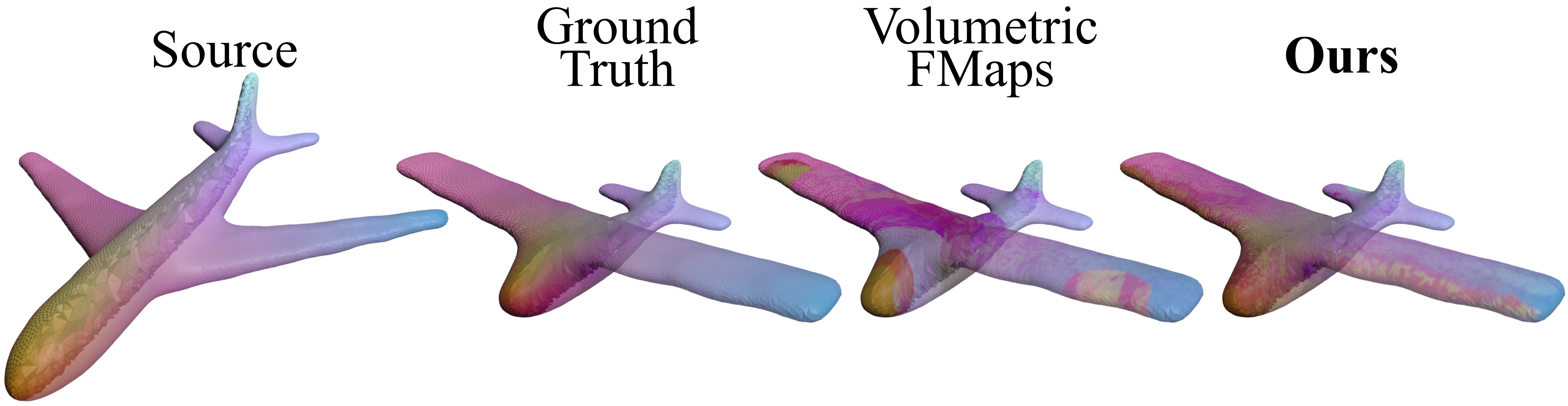}
\caption{Qualitative comparison of Ours vs Volumetric Functional Maps on a pair of non isometric tetrahedral meshes.}
\label{fig:volumetric}
\end{figure}
\end{minipage}% This must go next to `\end{minipage}`
\begin{minipage}{.42\textwidth}
\centering

%\vspace{-1cm}

\captionof{table}{
    Volumetric correspondence by accuracy and isometricity.
}
\label{table:volumetric}

\begin{tikzpicture}
\node[scale=0.72] {
\begin{tabular}{lccc}
    \toprule
    \multirow{2}{*}{\textbf{Mesh}} & \multirow{2}{*}{\textbf{Isometry} $\downarrow$}  & \multicolumn{2}{c}{\textbf{Accuracy} $\downarrow$} \\
    
     &  & {VolFmaps} & \textbf{Ours} \\
    \cmidrule(lr){1-1} \cmidrule(lr){2-2} \cmidrule(lr){3-4} 
    Octopus    & 0.0449 & \textbf{0.0151} & 0.0215 \\
    Cat        & 0.0561 & \textbf{0.0237} & 0.0422 \\
    Dino\_skel & 0.0851 & 0.0652 & \textbf{0.0426} \\
    Airplane   & 0.2207 & 0.1273 & \textbf{0.0942} \\
    \cmidrule(lr){1-1} \cmidrule(lr){2-2} \cmidrule(lr){3-4} 
    \textbf{Average} &  0.1017 & 0.0578 & \textbf{0.0501} \\
    \bottomrule
\end{tabular}
};
\end{tikzpicture}
\end{minipage}

% \giulio{CHECK NOTATION FORMULA}
%We quantify the isometry between two shapes as $
%E_{\mathrm{Iso}} = \mathbb{E}\left[\frac{|d_1 - d_2|}{d_1}\right] $, where $d_1$ and $d_2$ are normalized geodesic distances on source and target meshes. Lower values indicate stronger isometric preservation.

% \subsection{Notes on computational} % This can maybe go in the limitations
% Our method is significantly more expensive than alternatives. However, if we consider a collection of shapes, we obtain a universal representation that does not need additional optimization in computing pairwise correspondences.

\section{Conclusions and Limitations}
We presented \repname, a representation for maps between shapes based on Flow
Matching. By composing per-shape invertible flows through a shared anchor
distribution, \repname produces accurate, high-coverage correspondences that are
agnostic to the underlying geometric representation, supporting meshes, point
clouds, SDFs, and volumetric data. Its probabilistic formulation offers an
alternative to classical correspondence-based pipelines and provides a concrete
starting point for works integrating learned features with flow-based alignment.

\repname has three main limitations. First, symmetry disambiguation depends
entirely on the embedding: geodesic landmarks handle this well, but embeddings
lacking symmetry-breaking information (e.g., WKS) lead to degraded results.
Second, the reliance on consistent geodesic distances limits its effectiveness
on point clouds and makes it unsuitable for partial matching in its current
form. Third, the final KNN retrieval step (\cref{eq:knn}) makes the matching not
strictly bijective, as is standard in point-based methods, though the continuity
of both the GFE and the flow already supports a continuous solution. Extending
\repname to partial settings~\cite{xiehm2025echomatch,
bracha2024unsupervisedpartialshapecorrespondence}, enforcing full bijectivity,
and integrating learned smooth features are natural directions for future work.

\section*{Acknowledgements}
Lorenzo Olearo's PhD is funded by MUR under the grant ``Dipartimenti di
Eccellenza 2023-2027'' of the Department of Informatics, Systems and
Communication of the University of Milano-Bicocca, Italy. Daniele Baieri's
research is supported by a Humboldt PostDoc Fellowship from the Alexander von
Humboldt Foundation and previously by the PRIN project GEOPRIDE, 2022-NAZ-0115,
CUP H53D23003400001.
This work was supported by the NVIDIA Corporation with the RTX A5000 GPUs
granted through the Academic Hardware Grant Program to the University of
Milano-Bicocca for the project ``Learned representations for implicit binary
operations on real-world 2D-3D data.''

\bibliographystyle{splncs04}
\bibliography{bib}

% \twocolumn[{%
% \renewcommand\twocolumn[1][]{#1}%
% \maketitlesupplementary 
% \vspace{5pt}
% }]
% \maketitlesupplementary 

\newpage
\clearpage
\setcounter{page}{1}
\setcounter{section}{0}

\section{Additional \repname Details}

\label{sec:implementation_details}
\subsection{Implementation Details}
To implement the flows, we use the open-source \texttt{flow\_matching} library
\cite{lipman2024flow} and parametrize the vector field using a simple neural
architecture. Specifically, we employ a 256-dimensional, 6-layer MLP with Swish
activations~\cite{swish}, conditioned on both the input features \(x \in
\mathbb{R}^d\) and a scalar time variable \(t\). Before input to the MLP, the
features \(x\) and \(t\) are concatenated and processed through a Fourier
feature encoder with six frequencies, resulting in \((d+1)\,(6 \times 2) +
(d+1)\) features. A final linear layer projects the hidden representation back
to the original input dimension \(d\). Optimization is carried out similarly
as in ~\cite{zhang2024geometry}. We trained on a single NVIDIA A5000 GPU for
\(10^4\) epochs, sampling \(5 \cdot 10^4\) points per epoch uniformly from the
surface of each input shape. This setup requires less than 2 GB of GPU memory
and takes approximately 2 minutes per shape. Our implementation will be made
publicly available upon publication.

Comparing our setup to the one proposed in ~\cite{zhang2024geometry}, our choices drastically reduce computation time for training a single flow. While some design decisions in GeomDist can be advantageous for learning high-quality representations of highly detailed input shapes, our approach optimizes representations that are sufficiently accurate for our needs within minutes instead of the several hours reported in the original GeomDist implementation.

\subsection{Further Details on the Comparison Between FM and DDIM}
In this section, we show additional details highlighting the differences between
our method, built upon the FM framework, and the equivalent implementation with
DDIM, which is the basis of GeomDist~\cite{zhang2024geometry}.

\begin{figure}[!t]
\centering
\begin{overpic}[width=\columnwidth,grid=false,tics=5]{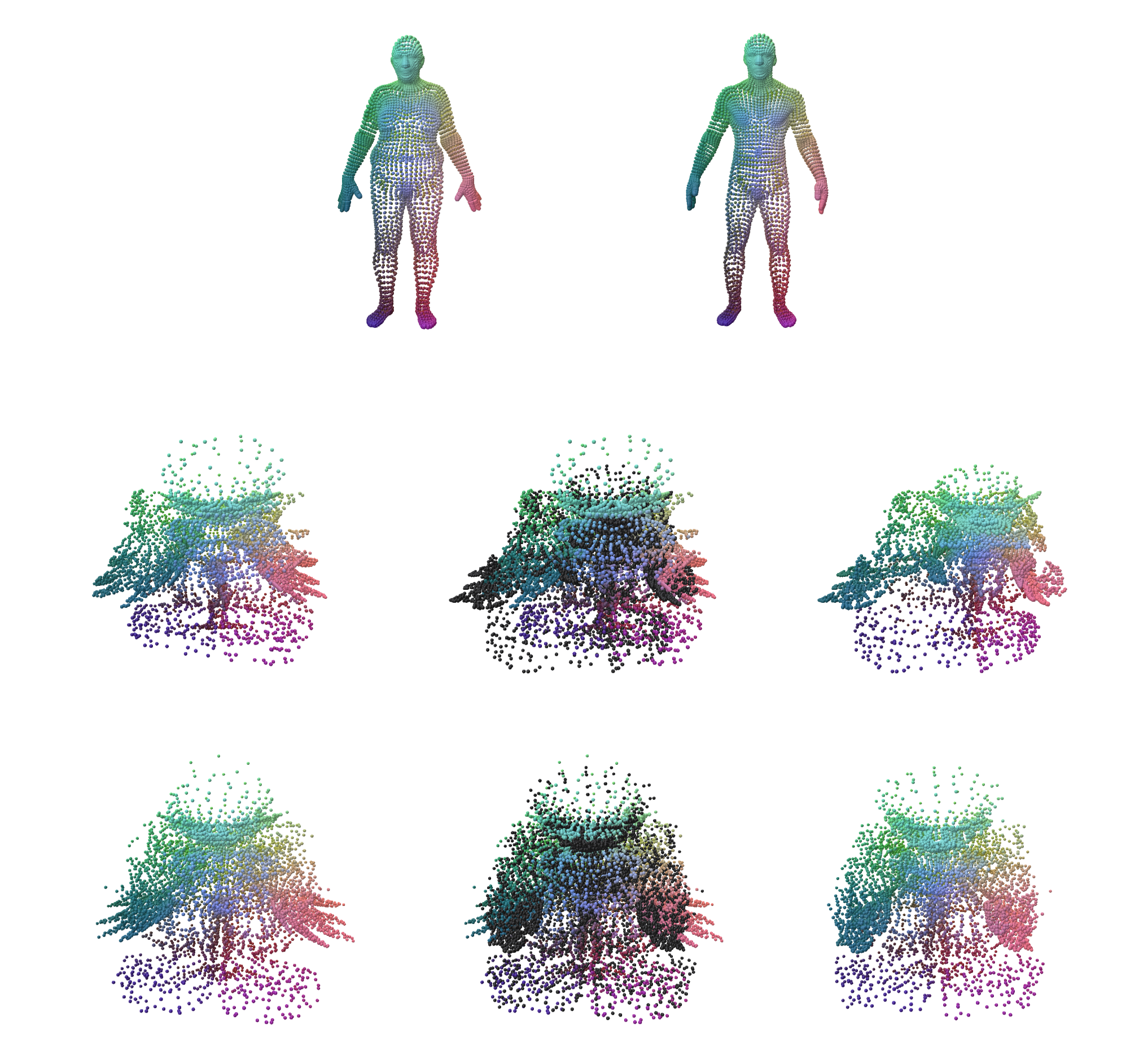}
    
    \put(34, 93){$S_1$}
    \put(65, 93){$S_2$}
    \put(42, 61){\tiny \textbf{DDIM Inverse Sampling}}
    \put(11, 58.5){\tiny $\left(\psi^{1}_{\text{DDIM}}\right)^{-1}(S_1)$}
    \put(75, 58.5){\tiny $\left(\psi^{2}_{\text{DDIM}}\right)^{-1}(S_2)$}

    \put(43, 31){\tiny \textbf{FM Inverse Sampling}}
    \put(12, 28.5){\tiny $\left(\psi^{1}_{\text{FM}}\right)^{-1}(S_1)$}
    \put(75, 28.5){\tiny $\left(\psi^{2}_{\text{FM}}\right)^{-1}(S_2)$}
    
\end{overpic}
\caption{ Visualization of differences during inverse sampling between Flow Matching (FM) and DDIM. The first row shows two shapes $S_1$ and $S_2$. The second
    row shows the result of the inverse sampling via DDIM for the two
    shapes. The column in the middle shows the aligned visualization to highlight the
    differences. The last row shows the corresponding inverse sampling results
    using Flow Matching. We observe a notable difference in the alignment of
    the two inverted shapes in the case of DDIM, which is quantitatively
    analyzed in \Cref{tab:comparison}. This difference highlights the improved
    accuracy and stable anchoring properties of FM compared to DDIM.
}
\label{fig:ours_vs_gd}
\end{figure}

\begin{table*}[t]
\centering
\caption{
    Full quantitative results for all $9$ cases of inter-representation
    shape matching. Best results per source-target pair are highlighted in
    \textbf{bold} and second best are \underline{underlined}.
}
\label{tab:inter-rep-full}
\resizebox{\textwidth}{!}{%
\begin{tabular}{@{}clcccc|cccc|cccc@{}}
\toprule
\multirow{2}{*}{Source Representation} & \multirow{2}{*}{Method} &
\multicolumn{4}{c|}{Target Representation: Mesh} &
\multicolumn{4}{c|}{Target Representation: PointCloud} &
\multicolumn{4}{c}{Target Representation: SDF} \\ \cmidrule(l){3-14}
& & Eucl. $\downarrow$ & Geod. $\downarrow$ & Dirich. $\downarrow$ & Cov. $\uparrow$
& Eucl. $\downarrow$ & Geod. $\downarrow$ & Dirich. $\downarrow$ & Cov. $\uparrow$
& Eucl. $\downarrow$ & Geod. $\downarrow$ & Dirich. $\downarrow$ & Cov. $\uparrow$ \\ 
\midrule

% ================= SOURCE: MESH =====================
\multirow{5}{*}{\parbox{2cm}{\centering Mesh}} 
& KNN                                                        & 0.0454 & 0.0434 & 0.0022 & 0.2452 & 0.1480 & 0.1404 & \underline{0.0016} & 0.0254 & 0.0543 & 0.0549 & \underline{0.0022} & 0.1842 \\
& OT~\cite{cuturi2013sinkhorndistanceslightspeedcomputation} & 0.0926 & 0.0856 & \textbf{0.0016} & 0.0890 & \underline{0.0747} & \underline{0.0722} & 0.0028 & 0.0472 & 0.0965 & 0.0910 & \textbf{0.0017} & 0.0945 \\
& NDP~\cite{li2022nonrigid}                                  & 0.0702 & 0.0886 & \underline{0.0020} & \underline{0.4893} & \textbf{0.0535} & \textbf{0.0643} & \textbf{0.0016} & \textbf{0.4828} & 0.0572 & 0.0712 & \textbf{0.0017} & \textbf{0.5114} \\
& \textsc{knn-in-gauss}                                      & \underline{0.0425} & \underline{0.0390} & 0.0241 & 0.3483 & 0.1121 & 0.1047 & 0.0395 & 0.2262 & \underline{0.0505} & \underline{0.0494} & 0.0247 & 0.3107 \\
& \textbf{Ours}                                              & \textbf{0.0289} & \textbf{0.0274} & 0.0066 & \textbf{0.5320} & 0.0957 & 0.0890 & 0.0184 & \underline{0.3761} & \textbf{0.0373} & \textbf{0.0375} & 0.0081 & \underline{0.4816} \\
\midrule

% ================= SOURCE: POINT CLOUD =====================
\multirow{5}{*}{\parbox{2cm}{\centering Point\\Cloud}}
& KNN                                                        & 0.1580          & 0.1508 & 0.0162 & 0.0129 & 0.0695 & 0.0685 & 0.0034 & 0.1540 & 0.1533 & 0.1465 & 0.0153 & 0.0117 \\
& OT~\cite{cuturi2013sinkhorndistanceslightspeedcomputation} & 0.1662          & 0.1556 & 0.0145 & 0.0339 & 0.0836 & 0.0802 & \underline{0.0025} & 0.1156 & 0.1629 & 0.1533 & 0.0142 & 0.0378 \\
& NDP~\cite{li2022nonrigid}                                  & \textbf{0.0547} & \textbf{0.0650} & \textbf{0.0017} & \textbf{0.4813} & \textbf{0.0530} & \textbf{0.0638} & \textbf{0.0018} & \textbf{0.4648} & \textbf{0.0578} & \underline{0.0715} & \textbf{0.0018} & \textbf{0.4753} \\
& \textsc{knn-in-gauss}                                      & 0.0805 & 0.0757 & 0.0300 & 0.1823 & 0.0650 & \underline{0.0639} & 0.0197 & 0.2333 & \underline{0.0645} & \textbf{0.0643} & 0.0123 & \underline{0.3995} \\
& \textbf{Ours}                                              & \underline{0.0695} & \underline{0.0680} & \underline{0.0129} & \underline{0.3725} & \underline{0.0645} & 0.0643 & 0.0123 & \underline{0.3995} & 0.0730 & 0.0716 & \underline{0.0107} & 0.3045 \\
\midrule

% ================= SOURCE: SDF =====================
\multirow{5}{*}{\parbox{2cm}{\centering SDF}}
& KNN                                                        & 0.0554 & 0.0554 & 0.0020 & 0.1697 & 0.1451 & 0.1387 & \textbf{0.0016} & 0.0254 & 0.0562 & 0.0569 & 0.0023 & 0.1744 \\
& OT~\cite{cuturi2013sinkhorndistanceslightspeedcomputation} & 0.0916 & 0.0873 & \underline{0.0019} & 0.0851 & \underline{0.0751} & \underline{0.0735} & \underline{0.0033} & 0.0489 & 0.0921 & 0.0881 & \underline{0.0018} & 0.0936 \\
& NDP~\cite{li2022nonrigid}                                  & 0.0530 & 0.0627 & \textbf{0.0015} & \textbf{0.5085} & \textbf{0.0535} & \textbf{0.0630} & \textbf{0.0016} & \textbf{0.4691} & 0.0525 & 0.0643 & \textbf{0.0016} & \textbf{0.5139} \\
& \textsc{knn-in-gauss}                                      & \underline{0.0475} & \underline{0.0465} & 0.0194 & 0.2706 & 0.1051 & 0.0979 & 0.0204 & 0.1612 & \underline{0.0472} & \underline{0.0469} & 0.0175 & 0.2784 \\
& \textbf{Ours}                                              & \textbf{0.0358} & \textbf{0.0361} & 0.0061 & \underline{0.4775} & 0.1004 & 0.0932 & 0.0102 & \underline{0.2680} & \textbf{0.0375} & \textbf{0.0385} & 0.0069 & \underline{0.4653} \\

\bottomrule
\end{tabular}
}
\end{table*}

\begin{table*}[t]
\centering
\caption{
    Average inference computational time (in seconds) and euclidean error for different methods on FAUST dataset.
}
\label{tab:computational_costs}
\resizebox{\textwidth}{!}{
\begin{tabular}{@{}l*{11}{c}@{}}
\toprule
Method                       & Ours   & Ours-ZO & Ours-NAM & fmap-wks & fmap   & fmap-ZO & fmap-NZO & \textsc{knn-in-gauss}  & ndp    & ot     & knn    \\ \midrule
time (s)        $\downarrow$ & 2.50   & 4.27    & 26.59    & 25.80    & 0.57   & 7.95    & 39.26    & 2.37                   & 3.36   & 1.47   & 0.05   \\
euclidean error $\downarrow$ & 0.0266 & 0.0093  & 0.0096   & 0.0270   & 0.0436 & 0.0097  & 0.0143   & 0.0393                 & 0.0616 & 0.0862 & 0.0608 \\ \bottomrule
\end{tabular}
}
\end{table*}

GeomDist (GD) is based on a Denoising Diffusion Implicit Model (DDIM) framework.
In this setting, the noisy data \( X_t \) evolves according to the discretized
update:
\begin{equation}
    X_t = X_0 + t \epsilon,
\end{equation}
where \( X_0 \) is the clean shape, \( \epsilon \sim \mathcal{N}(0,I) \) is
Gaussian noise. We note that the Gaussian noise only dominates the data
distribution asymptotically, as \( t \) grows large. This implies that the
latent isotropic Gaussian distribution is reached only for \( t \to \infty \),
as opposed to the finite-time interpolation guaranteed by Flow Matching. Thus,
GD's inversion map asymptotically approximates the Gaussian prior, leading to
deviations from the anchor distribution and hindering anchor space similarities
between different
shapes.%shared latent space properties between different shapes.

This difference underpins the improved accuracy and tighter distributional
alignment we observe in our flow-matching-based method.
Indeed, \repname's interpolation can be written as
\begin{equation}
    X_t = (1-t)X_0 + t X_1,
\end{equation}
where \( X_1 \) is the clean shape and \(X_0 \sim \mathcal{N}(0,I) \) is
Gaussian noise. We note that in this case, we have a clear notion of the time at
which we sample one of the two distributions, maintaining consistency between
different flows. In \cref{fig:ours_vs_gd}, we show a qualitative comparison
between two three-dimensional flows. Inverting the source and target flows to
their Gaussian-like base distributions reveals a subtle discrepancy in the DDIM
case. Under a Flow-Matching noise schedule, however, this discrepancy largely
disappears. For visualization purposes, the inverted target is shown in black
while the inverted source remains colored.

\section{Evaluation}

\begin{figure}[t]
    \centering
    \includegraphics[width=0.9\textwidth]{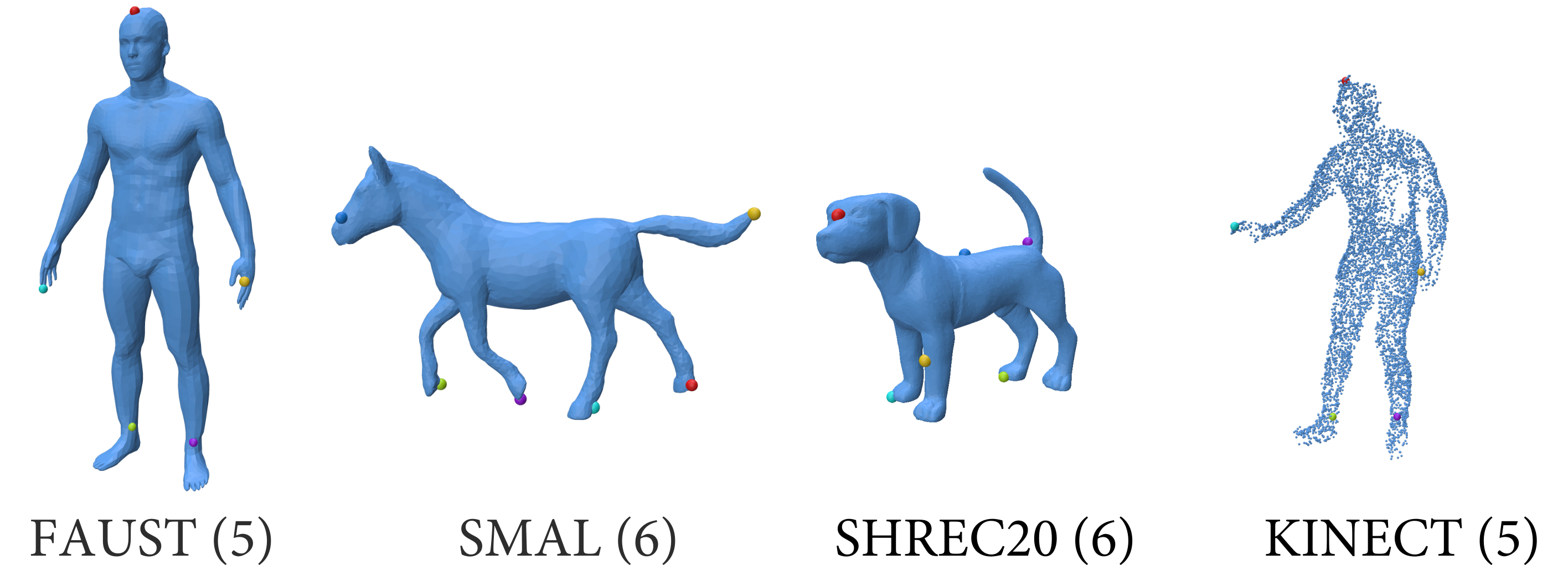}
    \caption{
        Landmarks used to compute geodesic embeddings (number per dataset in
        parentheses), hand-picked on one shape of the dataset and transferred to
        others via ground truth correspondences.
        Note that SHREC20 landmarks are selected from the small set of points common to all shapes in the datasets.
    }
    \label{fig:landmarks}
\end{figure}

\paragraph{Datasets.}
We evaluate our method on well-established datasets for non-rigid shape
matching:
\begin{itemize}
    \item \textbf{FAUST} (meshes): a dataset of near-isometric human subjects in
    different poses~\cite{bogo2014faust} in a 1:1 correspondence.
    \item \textbf{SMAL} (meshes): composed of different classes of quadrupeds in
    varying poses with strongly non-isometric deformations~\cite{Zuffi2017}; we
    use the remeshed version of the dataset~\cite{cao23unsup}, where shapes have
    different discretizations.
    \item \textbf{SHREC20} (meshes): 14 animal shapes including highly
    non-isometric pairs and shapes with missing parts~\cite{shrec20}. No dense
    registration is provided, each pair is annotated with sparse set of
    landmarks; we consider all the pairs of the $5$ test set and evaluate on the
    common landmarks between each shape pair.
    \item \textbf{KINECT} (point clouds): Kinect acquisitions of 15 human shapes
    interacting with objects from the BEHAVE
    dataset~\cite{bhatnagar2022behavedatasetmethodtracking} with SMPL
    registrations, introduced in~\cite{vigano2025nam}.
\end{itemize}
Landmarks are placed manually at shape protrusions: 5 for humans and 6 for
animals. For SHREC20 specifically, 6 points are selected from the pool of common
points shared across all shapes, with evaluation performed on the remaining
common point pairs. ~\cref{fig:landmarks} illustrates the landmarks on a
representative shape from each dataset.

\paragraph{Baselines.}
For Functional Maps-based baselines, we used implementations from the GeomFuM
library~\cite{vigano2025geomfum}. For NDP~\cite{li2022nonrigid}, we adopted the
original source code. For volumetric
FMaps~\cite{maggioli2025volumetricfunctionalmaps}, we considered the original
MATLAB source code. For FMaps and FMaps-WKS, maps were computed using 20 basis
functions. For the WKS variant, we leveraged 200 basis functions to compute
descriptors, using both landmark-based and global descriptors as standard.
ZoomOut and NAM~\cite{vigano2025nam} were implemented by performing spectral
upsampling from 20 to 120 basis functions with steps of 5 basis functions.
Optimal Transport (OT)
computations were performed using the Sinkhorn relaxation method~\cite{cuturi2013sinkhorndistanceslightspeedcomputation} available via
the PythonOptimalTransport library~\cite{flamary2021pot}.

\label{sec:evaluation_metrics}
\subsection{Evaluation Metrics}
In this Section, we provide details on how we computed all evaluation metrics
employed throughout the paper.
The Euclidean error is computed as: 
\begin{equation}
E_{\text{eucl}} = \frac{1}{|S_1|} \sum_{x \in S_1} \frac{\|T_{12}(x) - T_{\text{gt}}(x)\|_2}{D_{\text{eucl}}}\
\end{equation}
where $D_{\text{eucl}} = \max_{u,v\in S_2} \lVert u-v\rVert_2$.
Similarly, the geodesic error is defined as:
\begin{equation}
E_{\text{geod}} = \frac{1}{|S_1|} \sum_{x \in S_1} \frac{d(T_{12}(x), T_{\text{gt}}(x))}{D}
\end{equation}
where $d$ is the geodesic distance function on $S_2$, $T_{12}$ is the
estimated correspondence map, $T_{\text{gt}}$ is the ground truth map, and $D =
\max_{u,v\in S_1} d(u, v)$ is the shape diameter of $S_2$. 

The Dirichlet energy of the correspondence map is computed as: 
\begin{equation}
E_{\text{Dir}} = \text{trace}(T_{12}(X_1)^\top \bm{\Lambda} T_{12}(X_1) )
\end{equation}
where $\bm{\Lambda}$ and $X_1$ are respectively the Laplacian matrix and the vertex coordinates matrix
 associated with the source shape $S_1$. 

Coverage measures the fraction of the target shape \emph{covered} by the mapping, 
Importantly, we normalize the this measure by the cardinality ratio $|S_2|/|S_1|$ to
ensure comparability across shapes with varying numbers of vertices:
\begin{equation}
\text{Cov} = \frac{|\{ y \in S_2: \exists x \in S_1, T_{12}(x) = y \}|}{|S_2|} \cdot \frac{|S_2|}{|S_1|}
\end{equation}

The isometry metric evaluates the relative difference between pairwise geodesic
distances on tetrahedra that are in ground truth correspondence as:
\begin{equation}
E_{\text{iso}} = \sum_{x,y} \frac{| d_1(x,y) - d_2(T_{\text{gt}}(x), T_{\text{gt}}(y)) |}{d_1(x,y)}
\end{equation}
where $d_1$ and $d_2$ are geodesic distances on the source and target shapes
respectively.

\label{sec:boxplots}
\subsection{Evaluation Metrics Distribution}
\begin{figure}[b]
    \centering
    \includegraphics[width=\columnwidth]{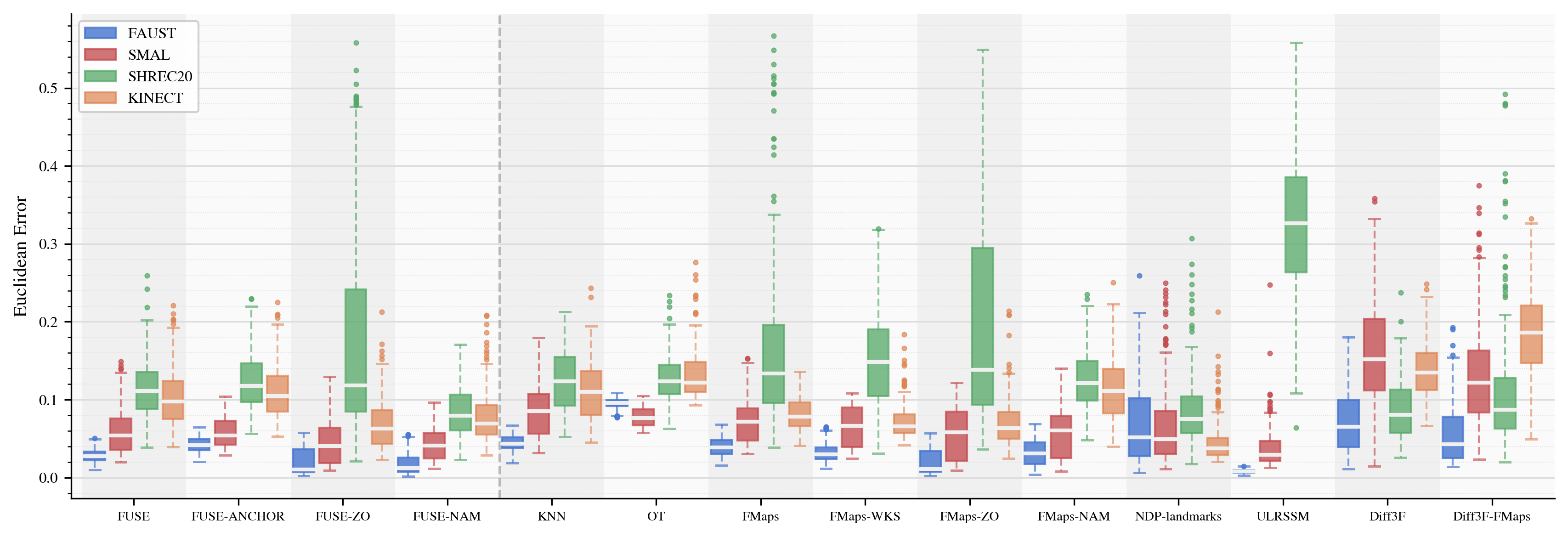}
    \includegraphics[width=\columnwidth]{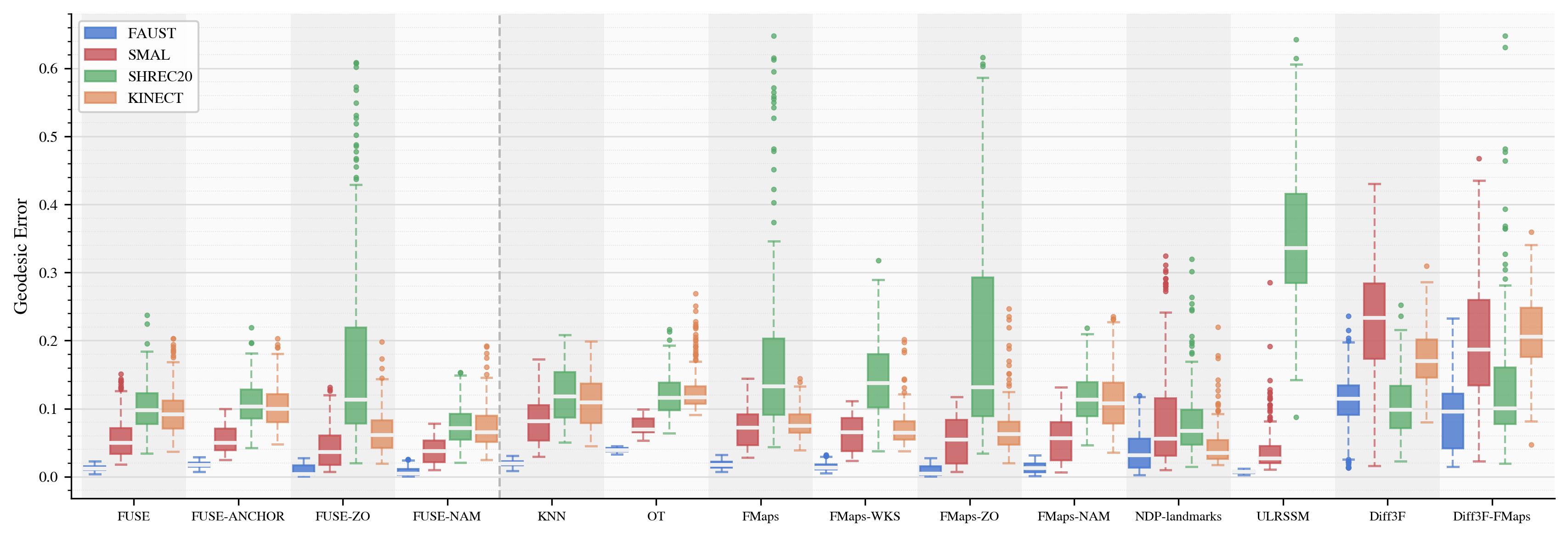}
    \includegraphics[width=\columnwidth]{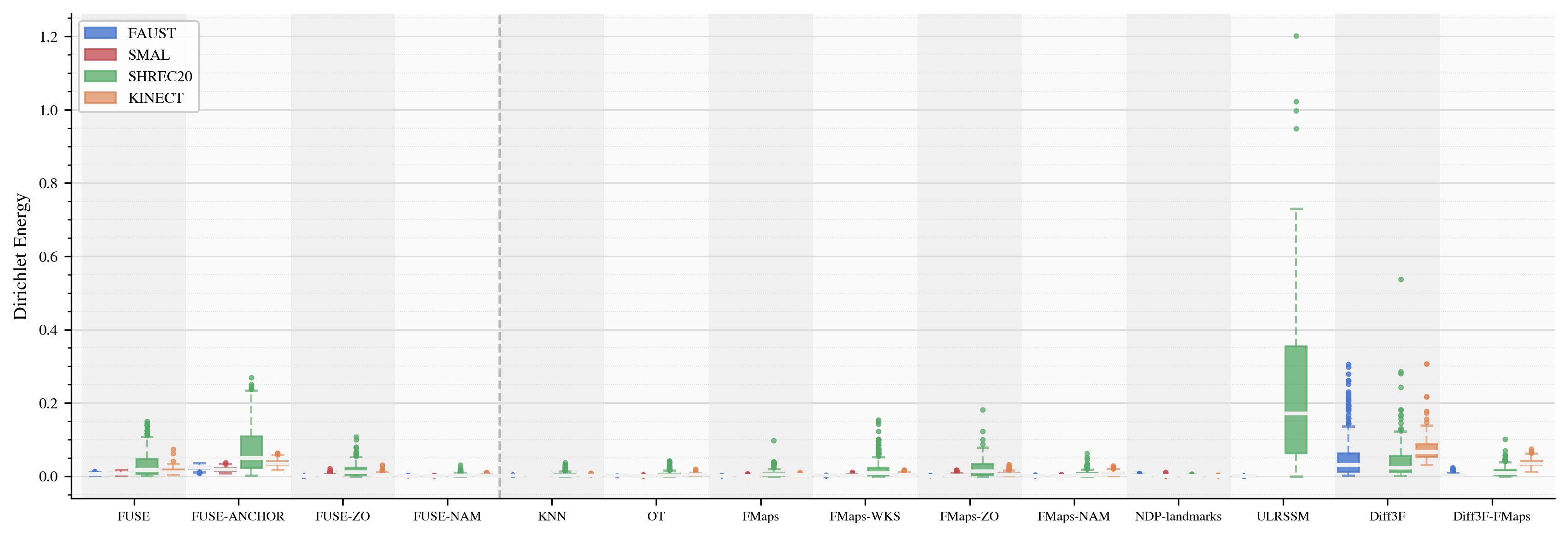}
    \includegraphics[width=\columnwidth]{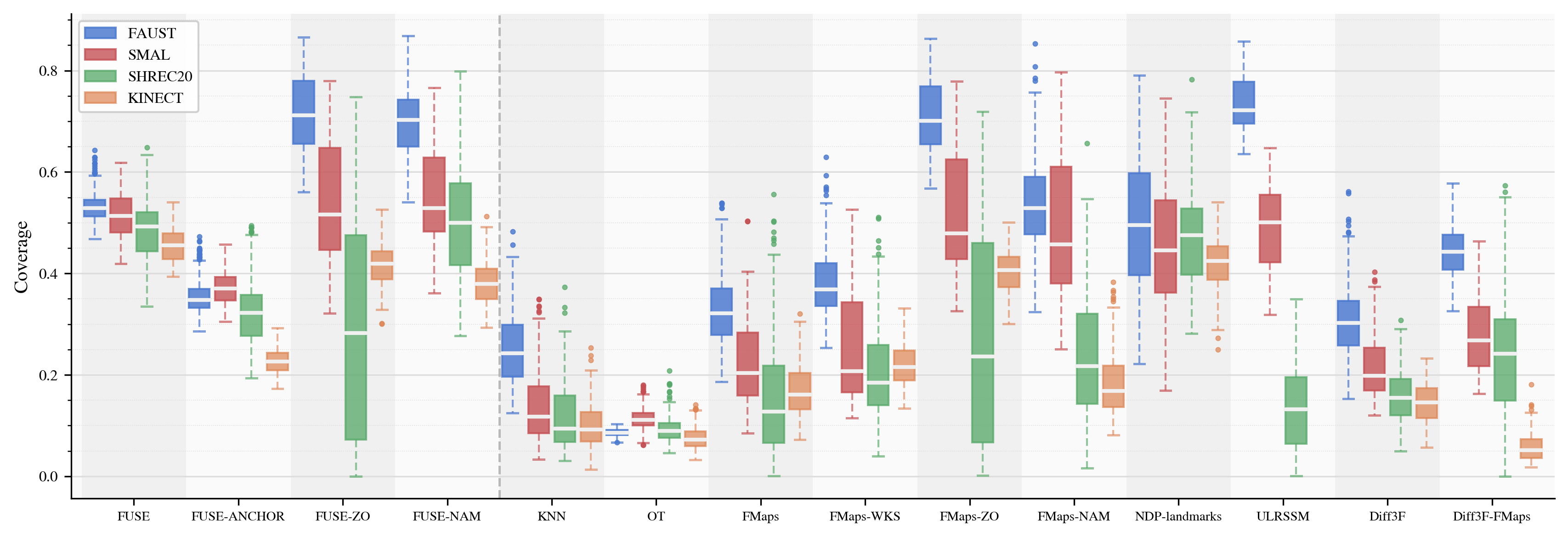}
    \caption{
        Distribution of Euclidean and Geodesic error, Dirichlet energy, and
        Coverage for each method on FAUST, SMAL, SHREC20, and KINECT. Each box
        spans the inter quartile range; the central line is the median while
        individual points are outliers. Methods left of the dashed line are
        ours.
    }
    \label{fig:boxplots}
\end{figure}

For a better visualization of the numerical results reported
in~\cref{tab:mesh-mesh-results} of the main paper, we provide
in~\cref{fig:boxplots} the distribution of per-pair Euclidean error, Geodesic
error, Dirichlet energy, and Coverage for each method across all four
benchmarks.

\label{sec:inter_representation_quantitative_results}
\subsection{Inter-Representation Quantitative Results}
We follow the setting described in \cref{sec:alternative-representations} to
evaluate our method in the context of inter-representation matching.
Out of all the baselines considered in \cref{sec:mesh-shape-matching}, we only take
into account those suitable for inter-representation matching,
namely KNN, Optimal Transport
(OT)~\cite{cuturi2013sinkhorndistanceslightspeedcomputation} and 
NDP~\cite{li2022nonrigid}, together with \repname and KNN in the Gaussian anchor
space (\textsc{knn-in-gauss}).

\cref{tab:inter-rep-full} reports the full quantitative results for all the $9$
inter-representation matching scenarios considered in
\cref{fig:multi_repr_match} of the main paper. Our method stands out in terms of
Euclidean and Geodesic errors when considering matching between
well-defined surfaces, such as meshes and SDFs. 
Consistent with the pointcloud-to-pointcloud matching results on the KINECT
dataset reported in \cref{tab:mesh-mesh-results}, \repname struggles when
matching to or from point clouds, where NDP~\cite{li2022nonrigid} instead
achieves the best results across all metrics, likely due to the shared
information at inference time. Furthermore, 
%NDP is a method specifically
%designed to handle point clouds, whereas
our method relies on accurate geodesic
distance computations and uniform sampling of the surface, both of which are
challenging to achieve on point clouds. Nonetheless, \repname still achieves the
second-best performance across most metrics in these scenarios.

\section{Smoothness}
% In this section, we include additional details and analysis on the necessity of smooth features for the functioning of \repname.
% First, we show an example of the effect of increasingly varying the smoothness of a feature on the performance of \repname. Secondly, we show how we can apply \repname to pretrained embedding by applying the right spectral filtering to obtain improved results.
% First, we define the $*$ operation, as the spectral filtering operation. Given the embedding $E$, we can obtain the spectral filter of the embedding $E*$ as 
In this section, we provide additional details and analysis on why smooth features are necessary for the functioning of the \repname.
Firstly, we demonstrate the effect that varying the smoothness of a feature has on the performance of \repname by providing an example. Secondly, we demonstrate that smoothing embeddings enables \repname to achieve better results even with high dimentional pretrained embeddings.
We define the $*$ operation as the spectral filtering operation. Given an embedding $E$, the spectral filter of $E*$ can be obtained as follows: 
\begin{equation}
    E* = \Phi_k\Phi_k^TM E
\end{equation}
where $\Phi_k$ is the $k$ truncated basis matrix obtained by the decomposition of the Laplace-Beltrami Operator of the input shape \cite{Levy08}. This operation is the same as that used in the main paper in Table \ref{tab:embedding-comparison}, with $k=10$.

\paragraph{Controlled experiment on smoothness.} 
To visualize the effect of smoothness on the effectiveness of \repname, we consider a pair of shapes from the FAUST dataset and compute their Wave Kernel Signature embeddings. We then perform spectral filtering while varying the number of eigenfunctions used. Increasing the number of eigenfunctions decreases the smoothness of the embedding and increases its Dirichlet energy. 

\begin{figure}[!t]
\centering
\includegraphics[width=\linewidth]{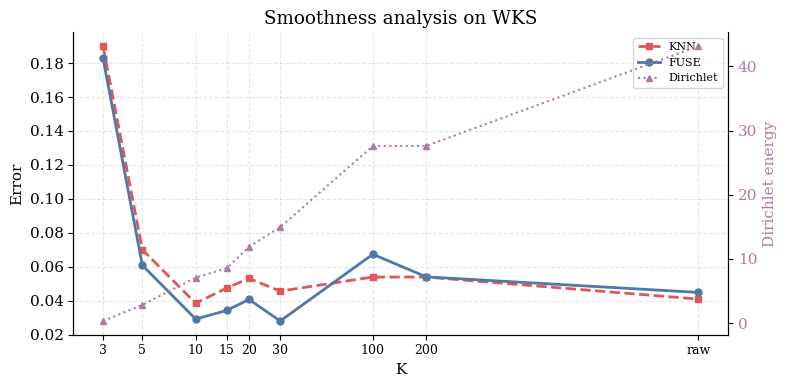}

\caption{Influence of the number of eigenfunctions used to perform low-pass filtering on a pair of shapes from FAUST using the WKS descriptors. Comparing KNN and \repname\ across different smoothing levels, we observe that \repname \ outperforms the baseline for smooth embeddings but underperforms for high-frequency embeddings. We also report the unnormalized Dirichlet energy for each feature used.}
\label{fig:wks_wksl_smoothing}

\end{figure}

Using these embeddings, we compute correspondences via KNN and \repname, which highlights a distinctive behavior of our method. In Figure \ref{fig:wks_wksl_smoothing}, we visualize this experiment. We observe that \repname improves the results when the embeddings are smooth, whereas for high-frequency embeddings, it degrades performance with respect to KNN. Interestingly, the combination of smoothing and \repname yields the overall best results.

\paragraph{Smoothed pretrained embeddings.}
\begin{wraptable}{r}{0.38\textwidth}
\centering
\vspace{-10pt}
\caption{Euclidean error on FAUST with original and modified embeddings computed using a pretrained feature extractor.}
\label{tab:faust_embedding_comparison}
\renewcommand{\arraystretch}{1.2}

\begin{tabular}{lcc}
\toprule
 & Original & * \\
\midrule
KNN        & 0.072 & 0.074 \\
\repname   & 0.082 & \textbf{0.068} \\
Dirichlet  & 0.1847 & 0.0016 \\
\bottomrule
\end{tabular}

\vspace{-5pt}
\end{wraptable}

We then consider pretrained embeddings obtained with the method of \cite{cao23unsup}. We show that applying the smoothing operation with $k=10$ improves the overall performance. 

In Table \ref{tab:faust_embedding_comparison}, we can clearly see that smoothing combined with \repname yields better results than the raw pretrained embeddings. This observation motivates future work on integrating \repname directly into the pretraining phase of embedding learning, and it is consistent with previously documented benefits of smoothness in learned representations \cite{attaiki2023understandingimprovingfeatureslearned}.

\label{sec:computational_costs}
\subsection{Computational Cost Analysis}
In Flow Matching, the vector field is parameterized using a neural network.
Consequently, beyond the computational cost of evaluating the network itself,
one must also take into account the numerical integration of the field to obtain
the actual flow. The computational cost of this process depends on both the
choice of solver and the number of integration steps employed.
In \cref{tab:integration_steps_ablation}, we present a comparison between Euler
and Midpoint solvers with varying numbers of integration steps, applied to both
the forward and inverse flow processes on the FAUST dataset. Notably, the
accuracy of the resulting maps is predominantly influenced by the precision of
the forward flow integration, whereas lower integration precision in the inverse
flow has a less pronounced impact.
As expected, the Midpoint solver with a larger number of integration steps
achieves better accuracy than naive Euler integration, albeit at the expense of
increased computational time. Conversely, employing fewer steps with Midpoint or
using a simpler solver such as Euler accelerates inference but results in a measurable reduction in mapping accuracy.

\label{sec:different_anchors}
\subsection{FUSE with Different Anchors}
A shape can be represented as a flow starting from any distribution from which
samples can be drawn. Naturally, the choice of this starting distribution can
influence the quality of the final maps obtained via the flow composition
employed in FUSE. In our experiments we used a standard Gaussian distribution as
the anchor, however, alternative choices are not only possible but may also be
better suited for other applications beyond the scope of this work.

The fundamental requirement for the anchor distribution in FUSE is robustness
with respect to flow inversion from the surface of the shape to itself. In
\cref{fig:different_anchors}, we provide a qualitative comparison of different
starting distributions in $\mathbb{R}^3$ for visualization purposes, while
\cref{tab:different_anchors} reports corresponding quantitative results on the
FAUST dataset with different multidimensional distributions. As illustrated in
\cref{fig:different_anchors}, sampling from the initial distribution $X_0$ and
subsequently applying the flow $\psi(X_0)$ produces a shape distribution that
deviates slightly from the target shape surface on which the flow was trained.
By performing a backward integration of the flow, $\psi(\psi(X_0))^{-1}$, we
recover an approximation of the identity map, with discrepancies coming from the
non-smoothness of the mapping from volume to surface of the inverted flow.

\begin{table}[t]
\centering
\caption{
    Results on the FAUST dataset when using different multidimensional anchor
    distributions for FUSE. We report the average euclidean error (Eucl.),
    geodesic error (Geod.), Dirichlet energy (Dirich.) errors, and coverage
    (Cov.). We highlight in \textbf{bold} choice of distribution used in the
    main paper.
}
\label{tab:different_anchors}

\begin{tabular}{@{}lcccc@{}}
\toprule
Anchor distribution & Eucl. $\downarrow$  & Geod. $\downarrow$  & Dirich. $\downarrow$ & Cov. $\uparrow$   \\ 
\midrule
\textbf{\emph{Standard Gaussian}} & \textbf{0.0289} & \textbf{0.0274} & \textbf{0.0066}  & 0.5320 \\
\emph{Unit Sphere Volume}         & 0.0292 & 0.0290 & 0.0094  & \textbf{0.5877} \\
\emph{Unit Volume}                & 0.0319 & 0.0318 & 0.0146  & 0.5840 \\
\emph{Unit Sphere Surface}        & 0.2654 & 0.2702 & 0.2477  & 0.3137 \\
\bottomrule
\end{tabular}

\end{table}

% \begin{table}[t]
% \centering
% \caption{
%     Results on the FAUST dataset when using different multidimensional anchor
%     distributions for FUSE. We report the average euclidean error (Eucl.),
%     geodesic error (Geod.), Dirichlet energy (Dirich.) errors, and coverage
%     (Cov.). We highlight in \textbf{bold} choice of distribution used in the
%     main paper.
% }
% \label{tab:different_anchors}
% \resizebox{\linewidth}{!}{
% \begin{tabular}{@{}lcccc@{}}
% \toprule
% Anchor distribution & Eucl. $\downarrow$  & Geod. $\downarrow$  & Dirich. $\downarrow$ & Cov. $\uparrow$   \\ \midrule
% \textbf{\emph{Standard Gaussian}} & \textbf{0.0289} & \textbf{0.0274} & \textbf{0.0066}  & 0.5320 \\
% \emph{Unit Sphere Volume}         & 0.0292 & 0.0290 & 0.0094  & 0.5877 \\
% \emph{Unit Volume} & 0.0319 & 0.0318 & 0.0146  &\textbf{0.5840} \\
% \emph{Unit Sphere Surface} & 0.2654 & 0.2702 & 0.2477  & 0.3137 \\ \bottomrule
% \end{tabular}
% }
% \end{table}

\input{Figures/multi-anchor}

In contrast, when sampling directly from the surface of the shape and applying
the inverse flow to project these points back to the anchor distribution, we
observe a more pronounced deviation from the expected anchor distribution. This
behavior is particularly evident in \cref{fig:shape-interpolation} and
\cref{fig:ours_vs_gd}, where the distributions obtained via the inverted flow do
not exhibit Gaussian-like characteristics. We attribute this phenomenon to the
fact that the sampled points occupy only a subset of the shape's distribution
parametrized by the flow; consequently, the resulting inverted distribution
corresponds to a subset of the anchor distribution.

To avoid smoothness issues during the inversion of a surface into a volume, one could consider using
an initial distribution that is naturally defined on a lower-dimensional manifold, such
as the surface of the unit sphere, thus having true diffeomorphic mappings
between the shape and the anchor. However, as shown in
\cref{tab:different_anchors} and \cref{fig:different_anchors}, this choice leads
to significantly worse results compared to volumetric anchors. This is likely
due to the challenges associated with accurately learning flows on
lower-dimensional manifolds embedded in higher-dimensional spaces, which can
result in instabilities during both training and inference.

\begin{table}[t]
\centering
\caption{
    Comparison between Euler and Midpoint solvers with different numbers of
    integration steps for the backward (bwd-steps) and forward (fwd-steps) flows on the
    FAUST dataset. We report average Euclidean error (Eucl.), geodesic error
    (Geod.), Dirichlet energy (Dir.), coverage (Cov.), and elapsed time in
    seconds (Time). The configuration used in the main paper is highlighted
    in \textbf{bold}.
}
\label{tab:integration_steps_ablation}

\begin{tabular}{@{}lcc*{5}{c}@{}}
\toprule
Solver & bwd-steps & fwd-steps &
Eucl. $\downarrow$ & Geod. $\downarrow$ & Dir. $\downarrow$ &
Cov. $\uparrow$ & Time (s) $\downarrow$ \\
\midrule
Midpoint & 10 & 10 & 0.0419 & 0.0428 & 0.0072 & 0.2933 & 0.41 \\
Midpoint & 10 & 64 & 0.0344 & 0.0349 & 0.0070 & 0.4259 & 1.43 \\
Midpoint & 64 & 10 & 0.0438 & 0.0450 & 0.0084 & 0.3527 & 1.46 \\
\textbf{Midpoint} & \textbf{64} & \textbf{64} &
\textbf{0.0289} & \textbf{0.0274} & \textbf{0.0066} &
\textbf{0.5320} & \textbf{2.50} \\
Midpoint & 32 & 32 & 0.0337 & 0.0340 & 0.0066 & 0.4692 & 1.26 \\
Midpoint & 16 & 16 & 0.0368 & 0.0377 & 0.0070 & 0.3669 & 0.64 \\
Euler & 64 & 64 & 0.0364 & 0.0360 & 0.0075 & 0.3945 & 1.30 \\
Euler & 10 & 64 & 0.0756 & 0.0700 & 0.0133 & 0.2479 & 0.74 \\
Euler & 64 & 10 & 0.0827 & 0.0778 & 0.0065 & 0.1954 & 0.75 \\
Euler & 32 & 32 & 0.0514 & 0.0496 & 0.0092 & 0.2791 & 0.64 \\
\bottomrule
\end{tabular}

\end{table}

% \begin{table}[t]
% \centering
% \caption{
%     Comparison between Euler and Midpoint solvers with different numbers of
%     integration steps for the backward (inversion) and forward flows on the
%     FAUST dataset. We report average Euclidean error (Eucl.), geodesic error
%     (Geod.), Dirichlet energy (Dir.), coverage (Cov.), and elapsed time in
%     seconds (Time). The configuration used in the main paper is highlighted
%     in \textbf{bold}.
% }
% \label{tab:integration_steps_ablation}
% \resizebox{\linewidth}{!}{
% \begin{tabular}{@{}lcc*{5}{c}@{}}
% \toprule
% Solver & bwd steps & fwd steps &
% Eucl. $\downarrow$ & Geod. $\downarrow$ & Dir. $\downarrow$ &
% Cov. $\uparrow$ & Time (s) $\downarrow$ \\
% \midrule
% Midpoint & 10 & 10 & 0.0419 & 0.0428 & 0.0072 & 0.2933 & 0.41 \\
% Midpoint & 10 & 64 & 0.0344 & 0.0349 & 0.0070 & 0.4259 & 1.43 \\
% Midpoint & 64 & 10 & 0.0438 & 0.0450 & 0.0084 & 0.3527 & 1.46 \\
% \textbf{Midpoint} & \textbf{64} & \textbf{64} &
% \textbf{0.0289} & \textbf{0.0274} & \textbf{0.0066} &
% \textbf{0.5320} & \textbf{2.50} \\
% Midpoint & 32 & 32 & 0.0337 & 0.0340 & 0.0066 & 0.4692 & 1.26 \\
% Midpoint & 16 & 16 & 0.0368 & 0.0377 & 0.0070 & 0.3669 & 0.64 \\
% Euler & 64 & 64 & 0.0364 & 0.0360 & 0.0075 & 0.3945 & 1.30 \\
% Euler & 10 & 64 & 0.0756 & 0.0700 & 0.0133 & 0.2479 & 0.74 \\
% Euler & 64 & 10 & 0.0827 & 0.0778 & 0.0065 & 0.1954 & 0.75 \\
% Euler & 32 & 32 & 0.0514 & 0.0496 & 0.0092 & 0.2791 & 0.64 \\
% \bottomrule
% \end{tabular}
% }
% \end{table}

\end{document}